# Autonomous programmable microscopic electronic lablets optimized with digital control.

Thomas Maeke[1], John McCaskill[1,4,5], Dominic Funke[2], Pierre Mayr[2], Abhishek Sharma[1], Uwe Tangen[1], Jürgen Oehm[3]

*[1] Microsystems Chemistry and Biomolecular Information Processing (BioMIP), Ruhr University Bochum, Germany*
*[2] Integrated Systems (IS), Ruhr University Bochum, Germany*
*[3] Analog Integrated Circuits (AIS), Ruhr University Bochum, Germany*
*[4] Research Center for Materials, Architectures, and Integration of Nanomembranes (MAIN, Chemnitz University of Technology, 09126 Chemnitz, Germany*
*[5] European Centre for Living Technology (ECLT), Ca' Bottacin, Dorsoduro 3911, Venice, 30123, Italy.*

## ABSTRACT

*Lablets* are autonomous microscopic particles with programmable CMOS electronics that can control electrokinetic phenomena and electrochemical reactions in solution via actuator and sensor microelectrodes. In this paper, we describe the design and fabrication of optimized singulated *lablets* (CMOS3) with dimensions 140x140x50µm carrying an integrated coplanar encapsulated supercapacitor as a rechargeable power supply. The *lablets* are designed to allow docking to one another or to a smart surface for interchange of energy, electronic information, and chemicals. The paper focusses on the digital and analog design of the lablets to allow significant programmable functionality in a microscopic footprint, including the control of autonomous actuation and sensing up to the level of being able to support a complete lablet self-reproduction life cycle, although experimentally this remains to be proven. The potential of lablets in autonomous sensing and control and for evolutionary experimentation are discussed.

## 1. Introduction

Lablets are autonomous microparticles with on-board electronics and power able to perform digitally programmed control of chemical systems via actuators and sensors. Autonomous electrochemical lablets were first conceived and explored in the MICREAgents project [1] [2] . This contrasts with earlier work, that linked chemical systems with electronic lab-on-a-chip systems, as a step towards programmable artificial cells [3], and electronic chemical cells [4], using fluorescence imaging feedback to replace on chip sensing. The lablet concept received inspiration from but it is distinct from ideas in smart dust [5], nanomorphic cells [6] and swarm computing [7]. It differs from smart dust in avoiding the scaling problems for wireless RF communication below 1 mm, especially in aqueous solution, by introducing a chemically inspired pairwise communication. This is based on mobile lablets docking to one another or to smart surfaces to exchange electronic or chemical information. Likewise, lablets were conceived independently of and follow a very different type of solution from that proposed in nanomorphic cells[6], with respect to communication architecture, containment strategy and powering. Although swarm computing is also conceivable with lablets, the primary initial focus in lablet design is to few particle systems, where docked lablets create a mutually accessible chemical space, separated from the surrounding medium in which their influence is large. Lablets are also designed to interact separately with a smart docking surface (dock) which has also been constructed and improved during the MICREAgents project. Currently this consists of an array of 128x128 pixels, each with 4 electrodes bearing electrochemical actuation and sensing capabilities [8].

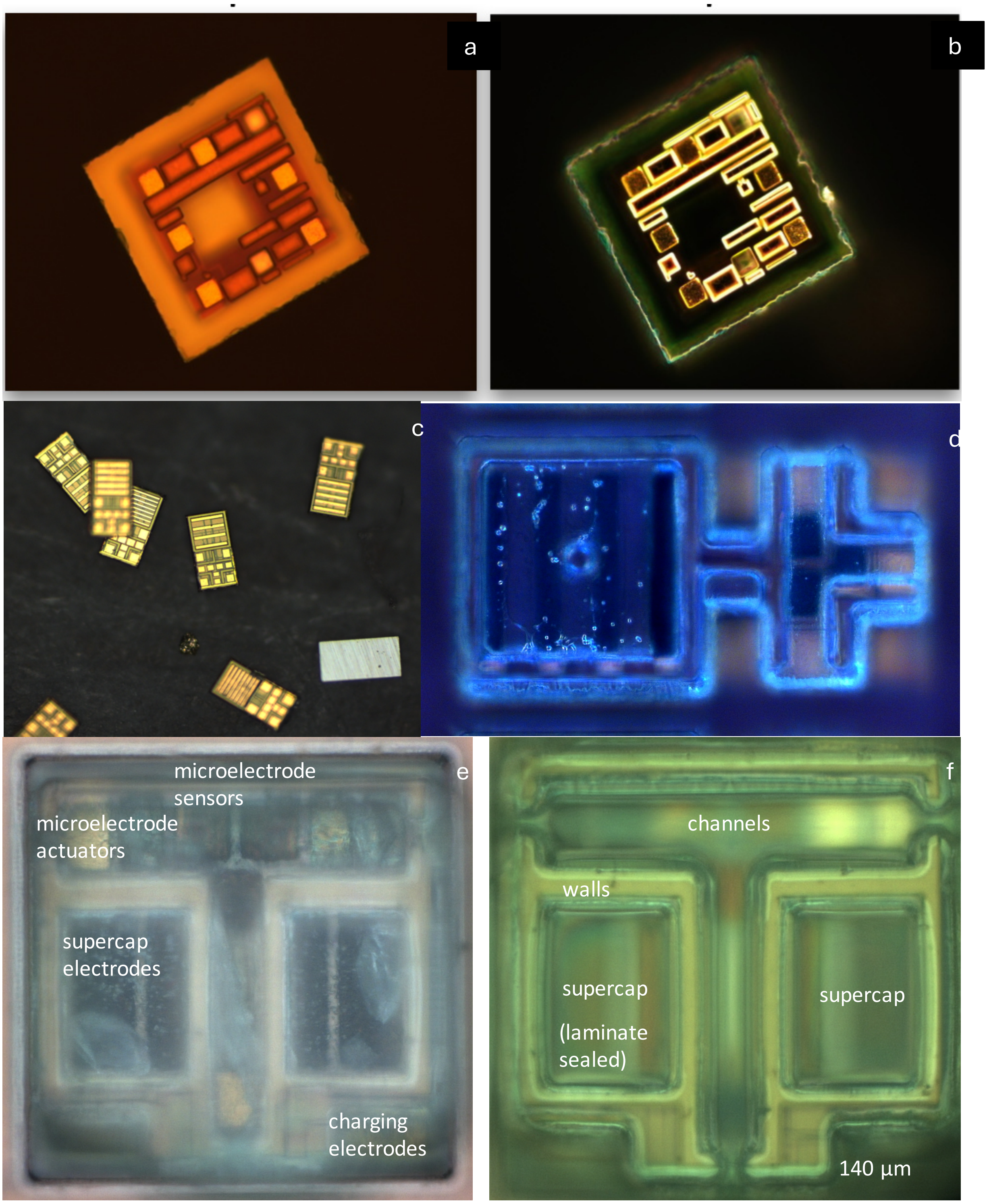

**Figure 1: Three generations of lablet design. Top row:** CMOS1 lablets without power (100x100µm) **a** brightfield, and **b** darkfield microscope images; **Middle row:** CMOS2 lablets with tandem supercap (100x200 µm) **c** without channel and chamber superstructures layers **d** complete, focus on covering lid layer; **Bottom row:** CMOS3 lablets with integrated supercap

(140x140 μm). The two images focus on the electrode plane (**e**) and microfluidic containment plane (i.e. walls**, f**).

Three generations of CMOS lablets have been produced, as depicted in **Fig. 1**.:

(i) CMOS1 lablets were constructed without a source of power to test the ability to fabricate and package electronics at this scale [9]. These lablets were 100x100x35 μm in size and had 8 gold-coated electrodes including 5 for powering and communication and a simple finite state machine with shift register programming to control actuation and communication.

(ii) CMOS2 lablets were made with a lateral coplanar supercap, rather than on the back side, for rechargeable power and electrode switching power surge buffering and on chip sensing with a size of 100x200x50 μm [10]. The chip is twice the size to accommodate the front side supercap.

(iii) CMOS3 lablets to be described in this work, with integrated supercaps and sensors and more flexible programming and control of actuation and sensing including support of locomotion control, with a size of 140x140x50μm. The increase in size compared with (i) and similar size to (ii) still stems from the use of a same side supercap (see below).

The digital design of minimal microcontrollers is a well-developed area of engineering, especially when resources are sufficient for an at least 4-bit microcontroller and sufficient program memory in the range of a kilobyte. After a careful evaluation of few-bit microcontroller architectures including 4-bit microcontrollers with Harvard architecture, efficient stack machine architectures [11], and 5-6 bit architectures designed for evolving self-reproducing automata [12], it was discerned that for a 100x100 μm footprint in the 2012-2015 available 180nm node CMOS technology, program length is prohibitive with on chip memory effectively limited to < 32 instructions (128 bits) in addition to controller logic and registers. Instead, a programmable state machine approach was adopted, with the state machine program encoded in a 58-bit shift register, taking advantage of the typically few phases of different programmable activity required to realize basic operations of lablets up to the level of autonomous life cycles.

In this work, we concentrate on the novel electronic architecture and verification of function of the lablets, while for fabrication we present only improvements compared with CMOS2 lablets [10]. The electronic architecture involves a modular separation of a digital program controller and an analog interface, the latter supporting slow clock [13], powering [14,15], communication, actuation and differential threshold sensing. This separation was essential for multi-team design progress, for placement autonomy and for combinatorial design variants. The analog design [16] is similar to that for CMOS2 lablet arrays [17] with some improvements that will be presented, whereas the digital design has been transformed to approach microcontroller flexibility and complexity within the limits of the gate counts available to the small lablet footprint. Although a move from 180nm to 135 nm node technology was considered, to increase the gate count, both financial and leakage current considerations as well as reuse of transistor designs dictated a more conservative way to complete the project. Lower leakage silicon on insulator technology was not available to us at these resolutions at the time.

The remaining structure of this paper is as follows. The following section presents the physical and functional design of the lablets, starting with a list of desired functionalities. The next two sections describe the analog and digital electronics modules and architecture of the lablets. Section five presents their physical variants and fabrication. Section six then demonstrates the electronic functionality, in agreement with simulation, for the CMOS3 lablets.

## 2. Physical and functional design of CMOS3 lablets

The physical and electronic design of the CMOS3 lablets (shown in 3D in **Fig. 2**) were targeted to support the following core functionalities:

(1) Autonomously clocked low power microcontroller electronics on board
(2) Integrated supercap for rechargeable power buffering
(3) Communication through solution to allow programming and data exchange
(4) Actuator electrodes switching under digital program control
(5) Sensor electrodes delivering threshold detection e.g. for pH and DNA
(6) Control of local ion and/or redox chemical concentrations via actuators

(7) Control of docking of lablets to a smart surface and/or one another
(8) Electronically triggered transfer of DNA between lablets
(9) Interaction with biosystems via electronically controlled drug release
(10) Autonomous control of lablet locomotion
(11) Execution of a complete lablet life cycle with combinatorial heredity
(12) Swarm collective dynamics and electronic-chemical translation

The first six of these are low-level functions that are required to support the second six as increasingly high-level functions. The physical design of lablets is reflected in decisions made to support 1-10, whereas the electronic architecture is designed to support autonomous program communication between lablets to enable 11 and 12. The experimental functional verification of the higher order functions will be the subject of separate work. The electronics design and functionality (1) required to support 2-12 will be presented in the following two sections.

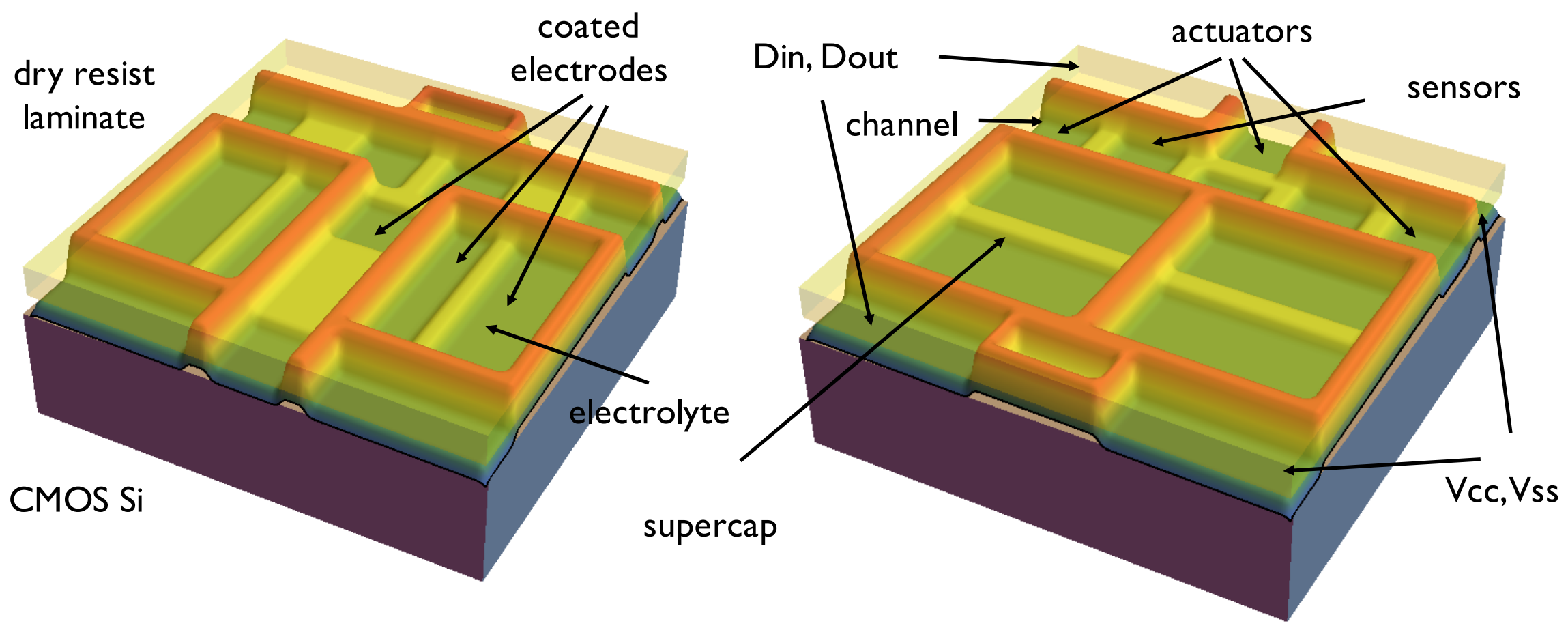

**Fig. 2 Geometric design of CMOS3 lablets.** Two of the three variants designed and built are shown. Lablet dimensions are 140x140x50μm. Half the surface area is taken up by the dual supercap structures, the other half by an optionally covered T-channel. The laminate enclosing structure is shown as transparent (simplified to a rectangular prism). The third variant implemented (see Fig. 9) is like the first, but with Din and Dout moved into the channel (top of T) to act as additional actors there.

The current dimensions of lablets is related to their supercap powering (2). The initial lablet target dimension was 100μm, which is at the edge of the microscopic domain and at the upper size limit for significant mobility [18], with later developments planned to take lablets down to at least 20μm. Docking (7) and higher order functions (9-12) require that lablets be microscopic for cell proximity and mobility, ideally approaching the size of cells. However, for power functionality (2), the surface area of the lablet had instead in this work to be doubled from 100μm square to 140μm square, to accommodate the supercapacitor (required area 100x100μm) along with actuators and sensors. In a 100x100 μm square lablet, the original intent (in 2012-13) was to place this same size supercap on the rear side, making use of TSVs (through silicon vias) to connect it with electronics. Technology for TSVs is available but the rear side supercap complicated the fabrication flow significantly and would have added additional risk to the fabrication of lablets. So in the second phase of lablet development it was first placed alongside the 100x100 μm lablet, resulting in the 200x100 μm shape of CMOS2 lablets, Fig. 1c,d. In the current work, the supercap is split into two parts operated in parallel to (i) minimize laminate sag (ii) add redundancy in case of fabrication defects (iii) keep diffusion lengths short for supercap power (iv) allow the placement of a symmetric T-channel on the lablet (v) pilot future potential use of two supercaps in series to raise the voltage (not currently supported by the electronics). The walls of the supercap were designed as 9μm wide (to ensure a reliable seal in the presence of mask alignment errors (± 2μm and microdefects) and 7-10μm high (to allow sufficient supporting electrolyte above the coated electrodes: 10μm being the minimal height required to support a supercap with 850mF/m$^2$ (85μF/cm$^2$) using 1M electrolyte). This aspect ratio <1.2 is well within the bounds of photolithography for wet or dry photoresists (see Section 5). In addition to the supercap, charging up the lablets requires two large power electrodes (Vcc/Vss also termed POWR1,2), open to the external solution in the non-encapsulated area. These are connected by diodes to the supercap and placed where they will have least influence on communication, actuators, and sensors.

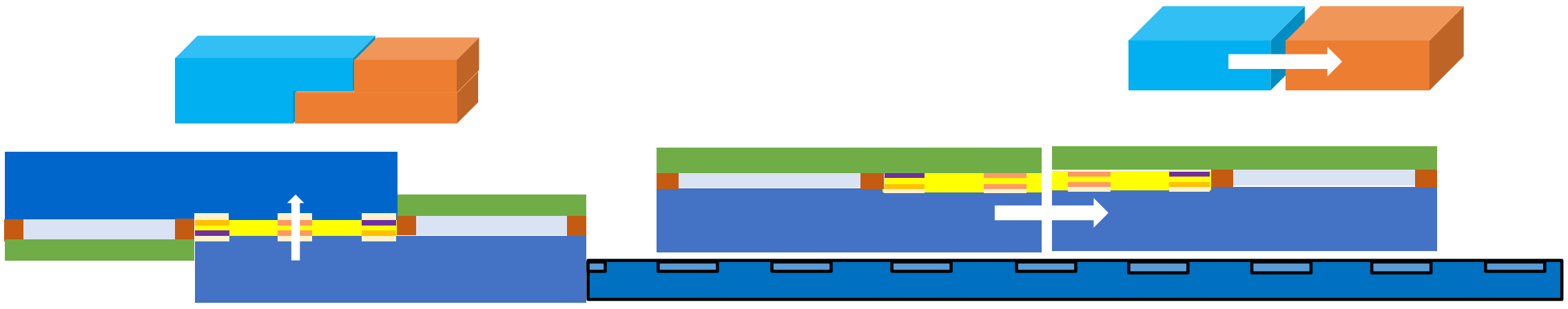

**Fig. 3 Docking and communication between lablets. Left:** Schematic showing facial docking of lablets. Facial docking between lablets is possible despite steric hindrance resulting from presence of supercap (encapsulated electrolyte shown in light blue). The semi-isolated fluid (yellow) supports strong local concentration changes via electrokinetic or electrochemical action of microelectrodes. Communication is vertical (white arrow). **Right:** Schematic showing alternative lateral docking of lablets on a smart substrate. Lateral communication of lablets with channels. Here communication with the nearest dock electrodes is vertical and over distances of 30µm, whereas lablet-lablet communication involves distances of 20-30 µm.

For communication (3), lablets need both to be able to control local concentrations in a protected portion of the solution as well as be able to communicate signals to the dock and other lablets. While the initial lablet concept [2] combined these two functions with a pair of lablets docking (7) facially, to enclose a layer of solution in a thin film between them and communicating with the other lablet through it, the presence of the supercap on the same side of the lablet makes the topographical realization of this more difficult (see **Fig. 3**). A version of the CMOS3 lablets without channels can still support this mode of lablet communication in a staggered docking of lablets. Of course, the original lablet concept with back side supercaps and free docking surfaces is preferable as soon as the technology for this is available. The alternative to the staggered docking mode of communication is lateral (see Fig. 3) , in which case the lablet can attain a controlled local volume to work with by itself using a T-channel, sealed on the top by the same laminate used for the supercap. T-channels are useful for flexible concentration control, rather than a straight channel or reservoir with a single pore [19-20]. Diffusion constants even for small ions in solution (~$10^{-9}$ $m^2/s$) mean that diffusional mixing times for dimensions of half a lablet are in the range of seconds, allowing microelectrodes to create strong local concentration changes in such channels. Lablets could communicate using their other actuators and sensors, but there are several advantages in using dedicated electrodes (DO,DI) for communication. To ensure greater orientation independence in powering, the two communication electrodes are also used for powering alongside the Vcc and Vss electrodes. This is accomplished using two diodes to connect each of the four electrodes to internal VCC and GND, a circuit design which also enables AC powering.

Sensors are placed symmetrically in the T-channel, with 2 different sensors and a reference electrode allowing with appropriate coatings pH or other chemicals e.g. DNA to be detected. Sensor and actuator areas are increased to the largest values possible in the channel in one variant, to optimize sensitivity. In another version, smaller electrodes are used to save power. The control and placement of actuators is presented in Section 4.

## 3. Analog design of CMOS3 lablet electronics

## Summary of analog electronic functionality of lablets and implementation in CMOS2

This section describes the analog sub-circuits that needed to be designed for lablet operation. We first present a summary of the analog circuitry in the precursor CMOS2 lablets [13,16], before turning to the enhancements in CMOS3.

Clock generator: The thyristor-based oscillator is a little-known concept that is extremely useful when designing ultra-low frequency, low area, low power oscillators. The time constant of thyristor-based oscillators is determined by the leakage current of MOS transistors and parasitic capacitance or small MIM (metal insulator metal) capacitors. The CMOS2 oscillator [13] yielded a frequency of 40-50 Hz on an area of approximately 35x15µm. Its power consumption was less than 300 pA for 1.5V and the frequency variation with the supply voltage, ranging from 1.8V down to 0.5V is less than 50%.

Sensor evaluation circuits: Integrating sensor circuitry into the lablets is challenging due to the lack of a stable voltage reference. Also, the requirements on current consumption of <<1nA are not met by conventional ADCs. The best solution found for small footprints restricts sensing to 1-bit comparator that can detect a threshold of the measured quantity *e.g.* pH. When one electrode is coated to be pH-sensitive (*e.g.* $IrO_2$) and one electrode is less sensitive to pH changes (*e.g.* Au), at a certain value of pH (which can be made adjustable in later work) the voltage difference between the two electrodes is 0V. This pH corresponds to the threshold, which is detected by the evaluation circuitry. To reduce power consumption, the comparator is duty cycled: The comparator is triggered at most once per clock cycle.

Power supply: The input rectifier of the CMOS1 lablets consists of one rectifying MOS diode. A special nwell potential regulation circuit was introduced in CMOS2, that automatically short circuits the nwell to either its drain or source contact, depending on which of these has the higher potential. The input rectifier is followed by one MOS diode that prevents current from flowing from the lablet's supercapacitor back into the solution. The input rectifier allows the lablet to be charged with AC or DC electric fields of both polarities. Driving electric fields for power can be produced from either the docking station or other devices.

Standard cell library: To cope with diverse digital logic (see section 3) a standard cell library that implements basic digital functions such as flip-flops and logic gates (AND, OR, Inverter etc.) is required. Such a standard cell library was provided by the CMOS manufacturer (IMEC/TSMC, Europractice). Unfortunately, the provided standard cell library did not meet the requirements in leakage current. A post synthesis simulation of a typical digital circuit yielded a leakage current of more than the total average power budget of a lablet of a 1-2 nA. Since doping modification was not available at the standard design input level, the only solution was to exchange the 1.8V transistors with 3.3V transistors. 3.3V transistors in the 180nm CMOS process employed have approximately double the threshold voltage, leading to a reduction of leakage current of approximately a factor of 1000 according to circuit simulation. Replacing 1.8V transistors by 3.3V transistors made it necessary to design an entire new library of standard cells. As standard cell libraries usually consist of several hundred cells, the redesign was limited to the most important and most used cells: AND2, AOI22, D-Flip-Flop, Inverter, NAND2, NAND3, OR2, Tri-State Buffer and several filler cells. With these cells, the leakage current of the digital circuits could be reduced by a factor of almost 1000 and is now almost negligible compared to other sub-circuits on the lablet. A drawback of this procedure is, that the circuit area increases by a factor of 1.5 since logic optimization is limited by the reduced set of available logic cell compared to a full library of standard cells.

Power on reset: After a given circuit is powered on, the state of its flip-flops is unpredictable. Depending on the function of a flip-flop, this may lead to unwanted behaviour of the circuit. Since resistors and capacitors could not be used to generate a delayed reset on the small lablet, a new power-on-reset circuit was developed for CMOS2, based on a voltage detector which signals when the lablet's supply voltage rises over approximately 300mV. A small digital circuit, after the voltage detector, ensures that the POR triggers only once after power on and that the clock signal is available when the reset signal is sent to the lablet circuitry.

Light block: The photoelectric effect in semiconductors may lead to malfunction of CMOS circuits due to unwanted charging and discharging of memory elements. Even more severe is the photo-induced occurrence of "latch-up". Latch-up is a state in which an unwanted low resistance connection between positive and negative supply voltage exists, causing the circuit to stop operating. To protect the lablets against photoelectric induced malfunctions, a layer of light blocking metal was included in all lablets. Most of the light block was placed in metal layer 4. If metal 4 was not available, metal layer 5 was used.

Supercap integration: For integration of the supercap various versions of supercap structures have been designed in top metal. The supercap structures were made with compatible interfaces to the lablets so that the different versions of lablets and supercaps can be combined easily.

### Optimization of analog electronic functionality of lablets in CMOS3

After analysis of the experimental results of the CMOS2 lablets, it was concluded, that the performance of the analog electronic circuitry can also be optimized in the CMOS3 production run. Three improvements to core functionality were made:

a) The design of a second oscillator to allow lablet versions with higher clock frequencies.
b) Adjustment of the sensor thresholds to allow multi-thresholding with Au/$IrO_2$.
c) The more extensive digital logic in the CMOS3 lablets (see section 3) and the new geometry of the lablets also required the placement of the analog interface to be reworked, along with the accommodation of sixth actuator option on one of the power (charging) electrodes, to allow a large area actuator to be used in lablet protocols.

In addition, an addressable lablet array was designed allowing the activation of a single lablet within the array or multiple lablets in parallel. This work, applicable to non-singulated lablet functionality, is reported in [21].

These improvements will now be discussed in more detail:

#### Design of a higher frequency oscillator

During experimentation with some versions of the CMOS2 lablets, it was found that the response of the lablet was too slow for optimal experimentation. For example, it was found in electrochemiluminescence experiments, that it took the state machine up two one minute until the first observations could be made at the actor electrodes of the lablet. Also the programming of the lablet took unreasonably long due to the low clock frequency. Therefore, based on the earlier version of the oscillator [13] a new version has been designed, featuring a 4 times higher oscillation frequency. The increase in oscillation frequency has been achieved by reducing the size of MOS capacitors.

### Optimization of sensor readout circuit

In CMOS2, a readout circuit for the chemical sensors on the lablet has been designed, which is depicted in **Fig. 4**. The CMOS2 readout circuit is able determine which of its two input signals is higher in potential. This information can then be processed by subsequent stages. In CMOS2 the sensor core was parametrized symmetrically, therefore the decision threshold is exactly 0V (neglecting manufacturing tolerances). This threshold can be shifted to positive or negative values by removing the symmetry within the sensor core. This has been done for the CMOS3 analog interface: One version of the CMOS3 analog interface contains two independent sensor readout circuits. Each of the two readout circuits has been parametrized to be unsymmetrically: one sensor will detect a positive voltage while the other will detect a negative voltage. The asymmetry has been achieved by changing width and length of the input transistors M1 and M2, which directly influence the time constants of the sensor core and thus the threshold voltage. The dependence of the transistor parameters on the threshold voltage is a non- linear and complicated function. Therefore, the necessary parameters have been determined by running a series of simulations and iteratively finding a better set of parameters.

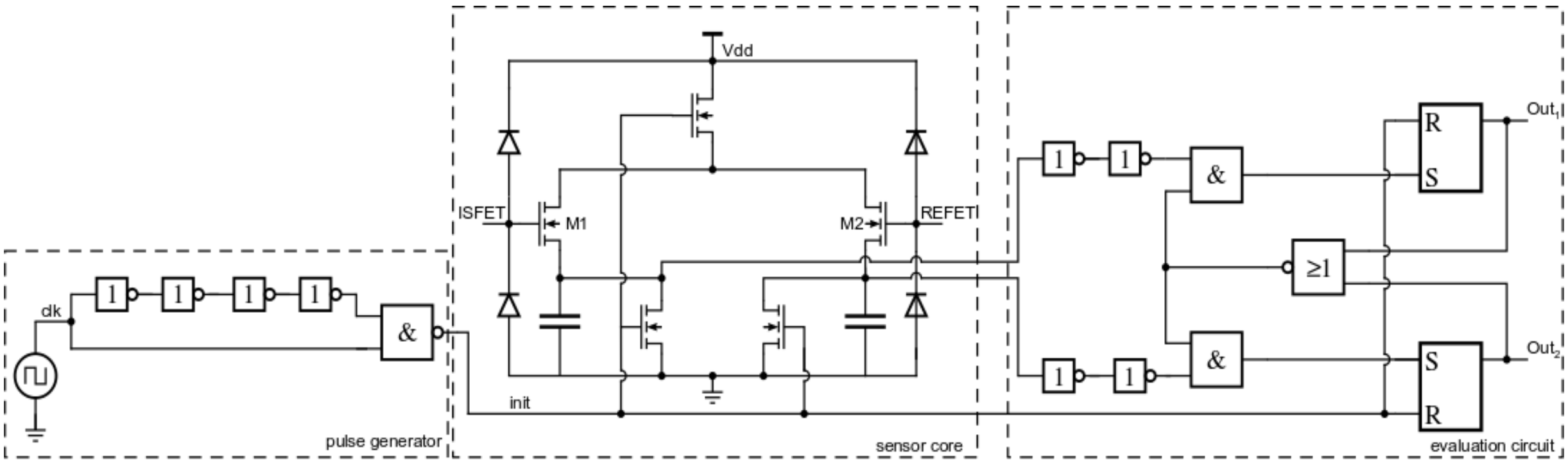

**Figure 4 Sensor readout circuit.** The circuit consists of a pulse generator, the sensor core and an evaluation circuit, that translates the output of the sensor core to binary signals. The coated chemical sensor electrodes are to be connected at nodes *ISFET* and *REFET*.

Simulation results of the two described sensor readout circuits in combination with the clock generator of the analog interface can be seen in **Error! Reference source not found. 5**. Sens<1> and Sens<0> are the outputs of the individual sensor readout circuits. In this simulation the voltage between node *ISFET* and *REFET* was continuously raised, as can be seen in the 2$^{nd}$ and 4$^{th}$ signal. The detection of surpassing the threshold voltage is communicated by raising the output signal from 0 to 1. As can be seen in the simulation, the threshold of Sens<1> is 85mV while the threshold of Sens<0> is -15mV. Since 1 pH unit corresponds to 59 eV, this potentially allows separate transitions in pH, close to 2 pH units apart, to be measured. These ultra-low power digital sensors, sampled only with the slow clock frequency, can be used to provide controller feedback to the digital logic in a wide variety of contexts. Not only a broad range of different electrochemical sensors, responding to different chemical concentration conditions, but also more sensitive detection of time dependent local kinetic potential differences e.g. from lablets signaling in the vicinity can be detected.

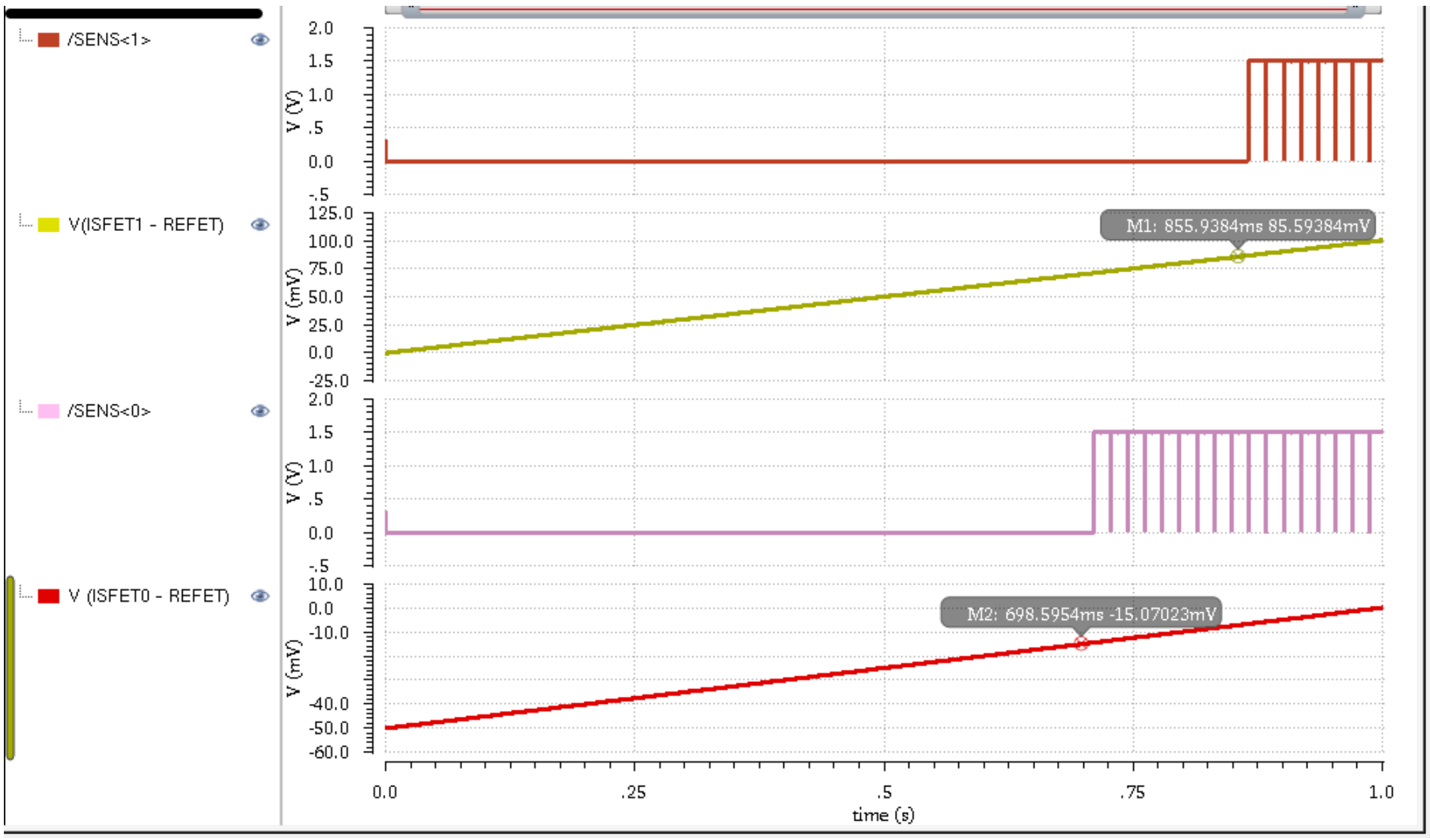

**Figure 5 Simulated two sensor response to different thresholds.** The sensors (1,2: -15mV, 85 mV) shown for increasing input voltage. The voltage of the REFET input was set to be 500mV.

## Adaptation of the analog interface to the new digital circuits.

The definition of an analog/digital interface, which is conserved for different variants, facilitates modular design development. The geometric location of the interfacing signals is also conserved, allowing separate partitioning and placement for digital and analog electronics. The CMOS3 analog/digital interface schematic connecting the signals is shown in **Fig. 6**.

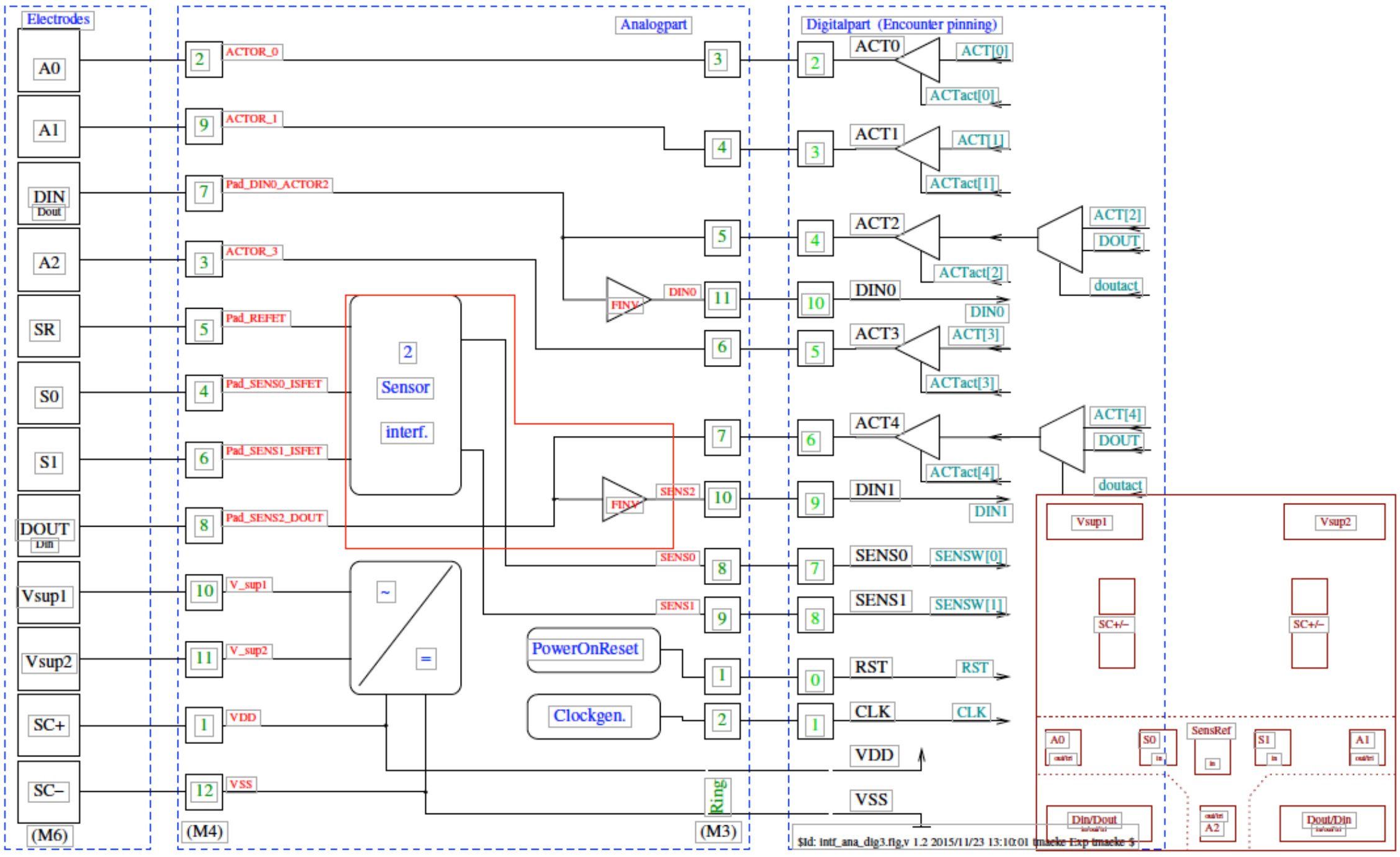

**Figure 6: Analog-digital interface for lablets.** 12 of the 14 lablet electrodes (the two IO pads not shown are those in parallel for the second supercap SC+, SC-) are shown in the block on the left: 3 dedicated actuators A0, A1, A2; three sensor pads S0, S1, SR; the digital IO DIN, DOUT (reversed in some lablets); the harvesting power supply pads $V_{sup1}$,$V_{sup2}$ and the supercap pads SC±. The analog and digital parts from the centre and right blocks, and an inset depicts a positioning on the lablet.

## 4. Digital design of CMOS3 lablet electronics

Firstly, three 4-bit microcontroller architectures were constructed and evaluated for their chip area requirements. The three designs chosen were custom designs, derived from three areas of microcontroller research (i) current minimal microcontroller practice with the Harvard architecture  (Thomas) (ii) a custom implementation of a stack machine microcontroller (John) and (iii) a microcontroller derived from considerations of minimal size for evolvability (Uwe). These three designs were only threshed out to the point where reasonable informed estimates of space requirements on the lablets could be made: it is estimated that additional features required for complete function may expand these requirements by up to about 20%.
Hardware synthesis was used to map these designs from a Verilog program implementation to CMOS via a customized library of standard cells. This showed that the simplified microcontroller architectures require a factor of at least two in integration density, which can only be considered at the next technology integration level (180nm -> 135nm). The three instruction sets are provided in the Appendix 1.

The memory required to store programs is a significant drain on resources, but a minimum length of program is necessary for even simple tasks to make the overhead of universal programmable operation worthwhile: 32 instructions seem too low a bound for practical use because a series of instructions are required to implement any proposed actuation change or iterative loop, with at least 64 really required (256 bit).

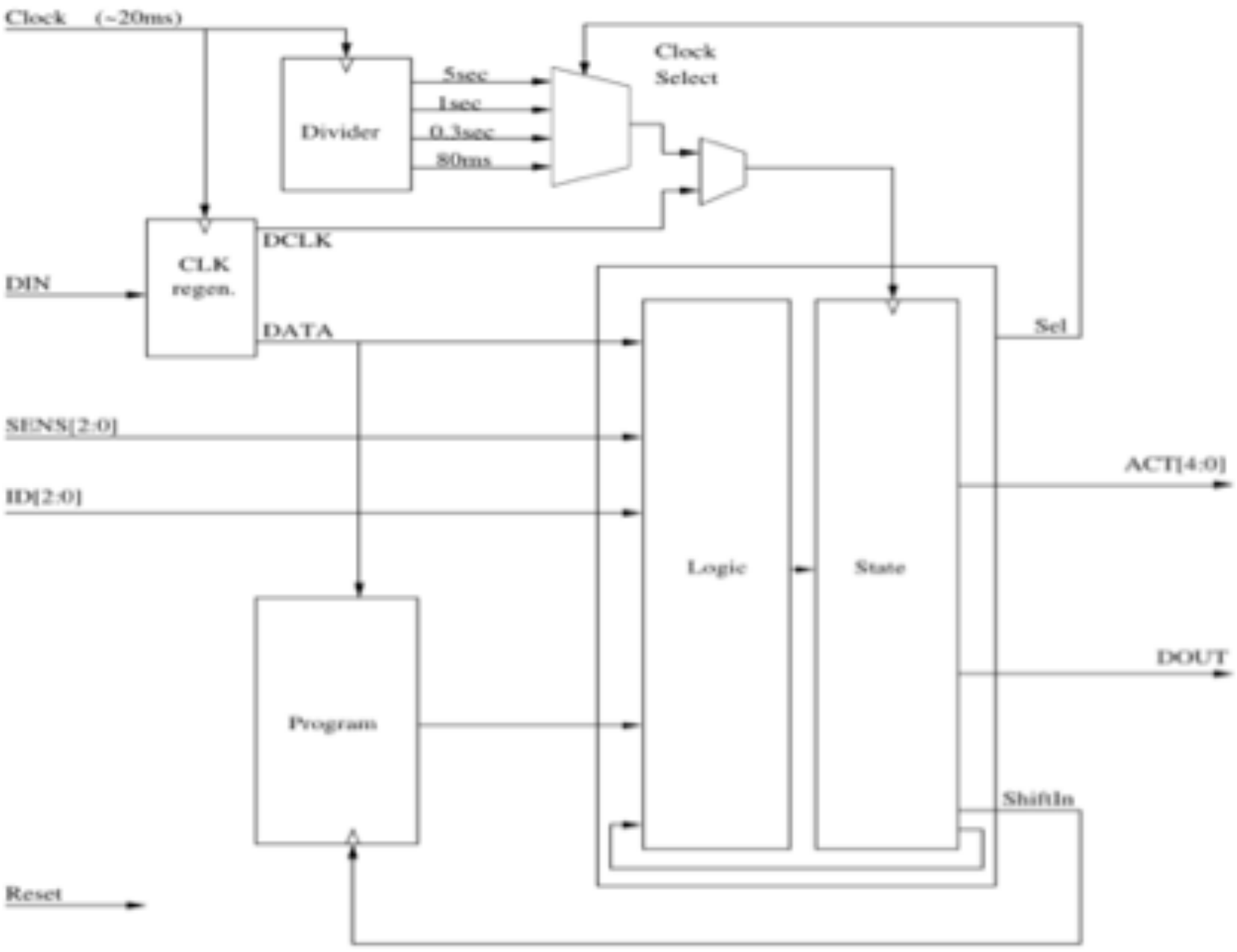

**Figure 7: Common architecture for  lablet digital design.** This semi-programmable finite state machine controller architecture used in the CMOS2 chips is employed with some modifications in the CMOS3 lablets. A programmable portion of the state memory can be loaded as a shift register and controls hard coded logic function in the rest of the state machine. Actuator and sensor signals in CMOS3 connect with the analog electronics as shown in Fig. 6.

In the previous generation CMOS2 lablets, the semi-loadable finite state machine architectures proposed for CMOS1, in which a subset of states can be loaded (like a program) using shift registers, were extended to provide more flexible programmable behaviour. Because of the strong restrictions on general programmability, a variety of different versions of semi-programmable lablets were designed to increase the overall range of lablet functionality achievable. They follow the overall principle shown in **Fig 7**.

The CMOS3 digital design presents an architectural compromise between general and specific purpose lablet design. Careful consideration of control loop requirements for typical chemical processes and the requirements for nested loop processing led to the symmetric design shown in **Fig. 8**. The lablet is in one of four operation modes or overall states (IDLE, PROG, SEND, RUN). These modes, add two communicating modes to the basic IDLE and RUN modes

(i) IDLE: Dormant mode, conserving power by not doing anything
(ii) PROG: being programmed (receiving data in communication)
(iii) SEND: transmitting information (sending data in communication)
(iv) RUN: running program mode: one of a finite set of up to three phase actuator/sensor feedback loops (which may include data collection)

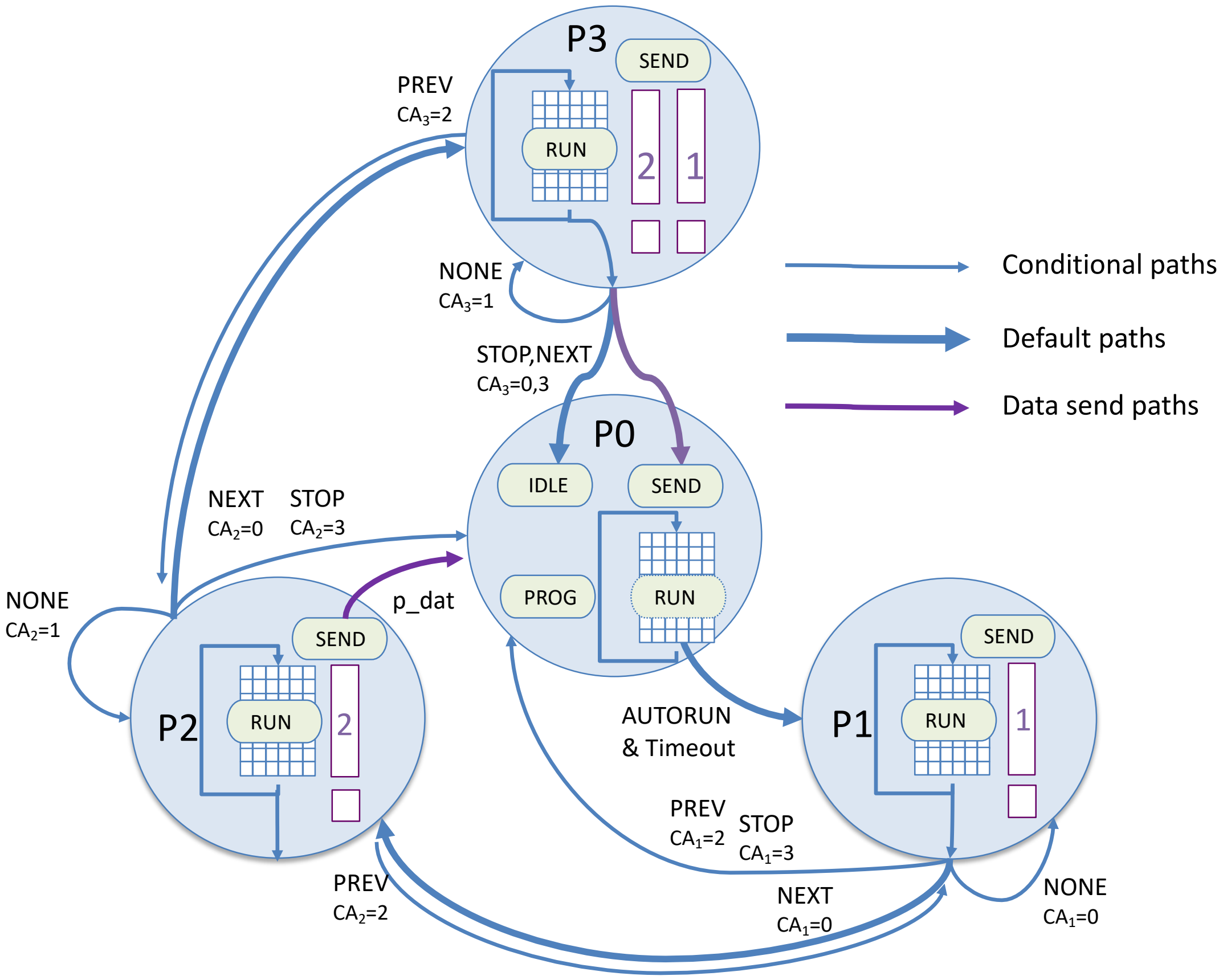

**Figure 8: Overall phase architecture of digital lablet controller.** There a 4 phases P0-P3 shown. Phase 0 is for lablet communication with three states: idle, prog and send. The other three phases are programmable running phases which execute a sequence of actuator patterns in a loop until a programmable condition is met. The third run phase can be used as the other two as an additional phase loop or as a recording memory for sensor or event timer data (programmable option TSDAT, see fig. 5), which can later be communicated as part of the (modified) program data. The programmable conditions also specify the next phase, so that programs may halt (ending at P0 idle) or iterate indefinitely at the phase level.

At the next level of detail, the programmable controller executes the lablet program (58 bits) according to the diagram shown in **Fig. 9**.

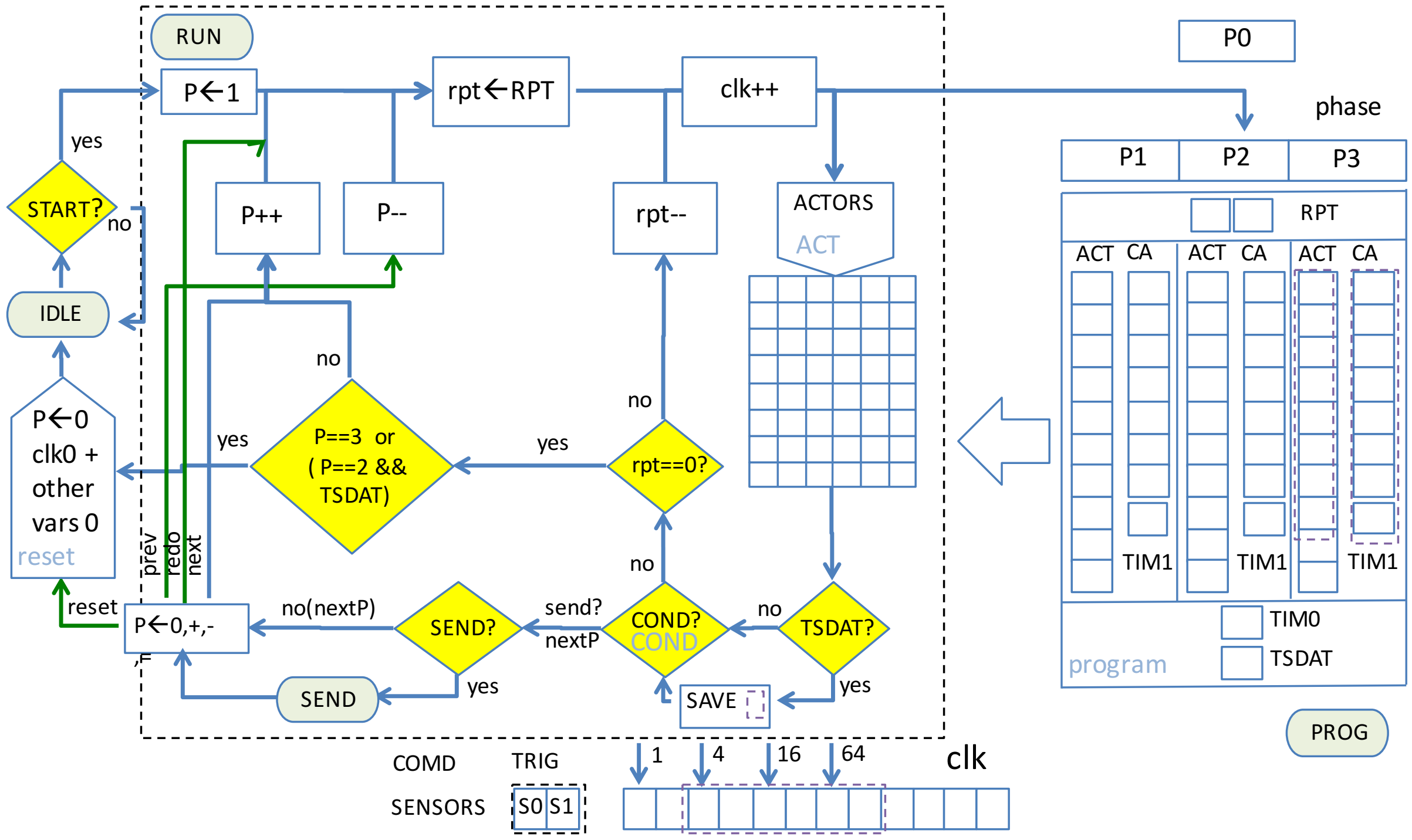

**Figure 9: Programmed execution of lablet controller in phase space.** The diagram is best understood as describing the transitional control a macro program phases P0-P3 and the detailed content of micro control loops. The innermost execution loop runs through 8 programmable actuation patterns for 6 electrodes (depicted as the 8x6 mesh). Time or sensor data is optionally saved and a programmable condition action determines if the second level repeat loop is continued down to zero from the programmed RPT count. If the condition (see Table 1) is fulfilled, either information is sent or not and the macro program proceeds to the programmed next phase (which may be the same phase).

Lablet program execution can be considered as a three-level architecture, with a top-level program pointer (P for the phase P0-P3) selecting the current macroinstruction from a program of length 4 (*cf.* Fig. 9), a second level counter (rpt) controlling the number of repeats of a fixed length (8) third level program pointer, cycling through a loop of 8 programmable microinstructions specified in each such phase (driving actuators according to programmable patterns ACT). The second level rpt counter is controlled with a programmable length (RPT) and exit condition (CA) which can depend on sensor or other events (trig, comd, see **Fig. 13**), resulting in a programmable multi-branch transition to the chosen next phase. The possible conditions are listed in Table 1. This architecture acknowledges the generic entry-level behaviour of lablets required as following a programmable sequence of transitions between defined periodic actuation patterns, repeated until certain conditions are fulfilled. Examples include the repeating of an electrochemical coating pattern on specific electrodes until a certain pH sensor signal is attained.

The maximum number of memory bits that can be packed into the lablet depends on the complexity of the control logic, but with the kind of control envisaged here, it is less than 60 bits for the available silicon in the CMOS3 sized lablets. Each of the three programmable phases (P1-P3) is specified by 18 bits of program memory, as depicted on the right of Fig. 9. Most of the program memory in each phase (17/18 bits) is used to specify a sequence of 8 actuation patterns for 6 actuator electrodes (10 bits, see below) and one of 128 conditional actions (4 bits encoding the condition and 3 bits the action). The conditions are listed in Table 1. The actions involve the choice of next phase (2 bits), and whether to send a data packet (SEND). The final memory bit in each phase is for the step time in that phase as described below.

A programmable clock operates at one of 4 periods which are successive 4x multiples of the slow clock fundamental (see section 2), specified by two bits of program information, one global and one specific to each phase (TIM1). This allows the program to specify both the overall processing unit time (x1, x4) and to choose a faster or slower repeat time step (x1, x16) in each phase resulting in one of four periods (x1, x4, x16, x64). For example, these would extend a 20Hz slow clock to actuator steps on the timescale of 50ms, 200ms, 800ms and 3.2s. Some of the lablets were fabrication with a slow clock running 10x faster at 200 Hz, see section 2, extending the range of possible time steps to almost a factor of 1000 (5ms to 3.2s).

| Code | Condition | Description |
|---|---|---|
| 0 | False | Default, no action |
| 1 | S0 ≥ Th0 | Sensor 0 |
| 2 | S1 ≥ Th1 | Sensor 1 |
| 3 | S0≥Th0 && S1≥Th1 | Both sensors |
| 4 | S0≥Th0 ≠ S1≥Th1 | Sensors differ |
| 5 | Trig | Trig cmd received |
| 6 | Trig && S0≥Th0 | Trig and sensor 0 |
| 7 | Trig && S1≥Th1 | Trig and sensor 1 |
| 8 | S0 < Th0 | Not sensor 0 |
| 9 | S1 < Th1 | Not sensor 1 |
| 10 | comd == 1 | Command rec'vd |
| 11 | comd == 0 | No command rec'vd |
| 12 | S0 != MemS0 [P] | Sensor 0 new val |
| 13 | S1 != MemS1 [P] | Sensor 1 new val |
| 14 | S0 != MemS0 [P] \|\| S1 != MemS1 [P] | New sensor value |
| 15 | S0≥Th0 && S1<Th1 | Sensor 0 only |

**Table 1. Programmable condition actions in each phase.** S0 and S1 are the two sensor values. Thresholds for sensors (difference to reference potential) Th0 and Th1 depend on lablet type. Trig is an external trigger signal, comd specifies whether an external command has been received, and MemS0/1 are the recorded binary sensor values in phase P. These remembered values (in 12-14) are values saved last in memory when TSDAT bit set. Arrayed lablets, in the final CMOS3 tapeout, used a Verilog option to restrict the conditions to 0-7, to save space and power.

The programming of actuation patterns presented a dilemma, because not only are different patterns needed in the different phases, but meaningful functionality requires controlled sequences of activation patterns. A sequence of 8 patterns was chosen as a trade-off between the logic cost of loop management and memory requirements (**Fig.10**). With binary 0,1, and tristate (Z or 2) values and 6 actuators there are $6^3$=216 different patterns (requiring 8 bits to specify) and specifying 8 such patterns require in general 64 bits, whereas only at most 10 bits can be afforded. A carefully optimized compromise solution is presented in **Fig. 10**. Its connection with real lablet electrodes for three lablet configurations is shown in **Fig. 11**.

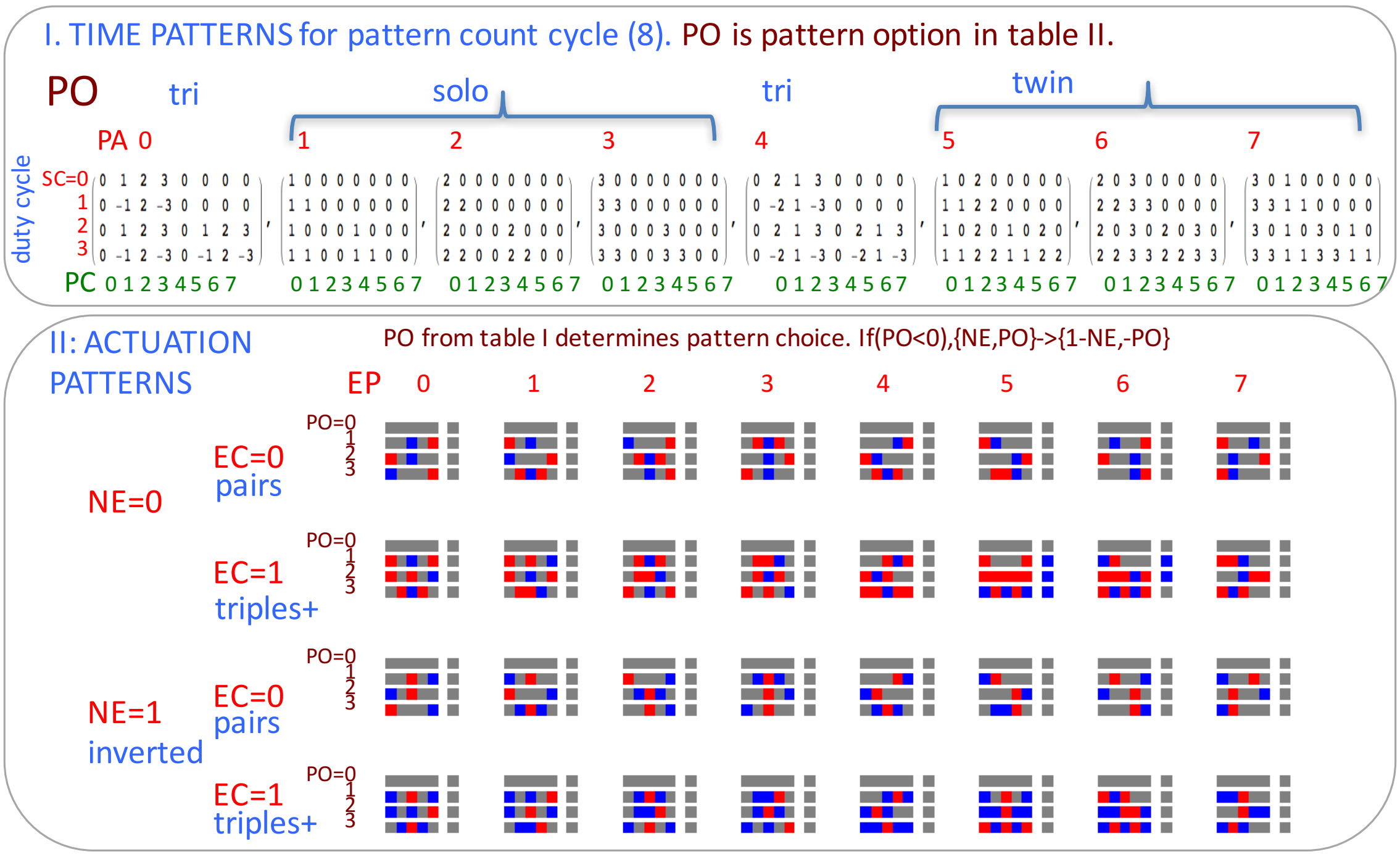

**Figure 10: Programmable lablet actuation pattern sequences.** The upper table gives the pattern choice PO for the 8 counts of the innermost clock (green PC in top figure) as a function of two program variables PA and SC. The pattern address variable PA determines the deployment of up to 3 different patterns during the 8 counts – either solo, twin or tri. The SC variable changes the saturation count for these patterns between light and heavy (equivalent to an increasing duty cycle for activation). The pattern variable PO can take four pattern values 0-3, with optional minus sign for pattern negation. The lower table shows the patterns employed as a function of three variables EP,EC, NE for the possible values of PO. Note that the absolute value of PO is used, but NE is inverted 0<>1 if PO was negative. The six actuator electrodes are A0, DO, A2 DI A1 and PWR2 as shown in fig. 11. Red is 1 and blue is 0.

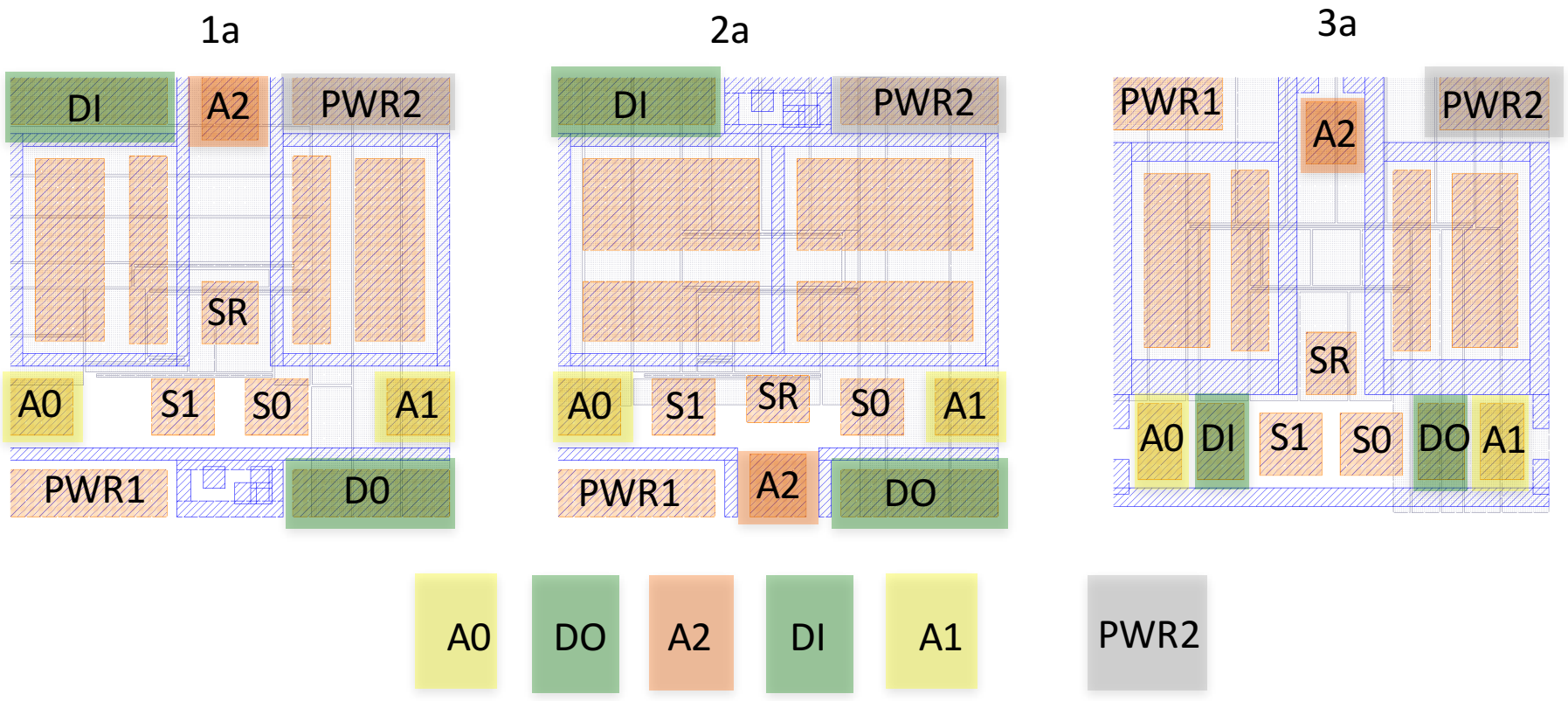

**Figure 11: Lablet actuators and connection with activation patterns for 3 types of CMOS lablets.** The common activation patterns shown at the bottom label the electrodes on the lablets and correspond to the patterns shown in Fig. 10. One of the PWR electrodes (for charging) can optionally be driven as an actuator, but only negatively. The three actuators A0-A2 are at the end of the channels on the lablet in all designs. The remaining two electrodes' data in and data out (DI, DO) can be used as actuators and are either deployed outside the lablet channels (1a,2a) or inside (3a). The sensor electrodes S0,S1 and sensor reference SR cannot be activated.

The coding of pattern sequences is computer generated and systematic, weighted towards simplicity both in the number of different patterns in the sequence and in the number of active electrodes in any single pattern. The 6 actuating electrodes on lablets have a certain symmetry that can be addressed in the following diagram: the nodes represent the 6 actuators and the edges pairs of electrodes that may be active at the same time. Half of the combinatorial resources are devoted to all pair activation patterns, see **Fig. 12** (note that single electrode activation will only be very weakly sustainable by leakage currents since a competed lablet circuit is necessary to sustain Faradaic reactions). Here the lablets are fully programmable. A sample of triple and higher order activation patterns deemed important make up the other half of the combinatorial repertoire.

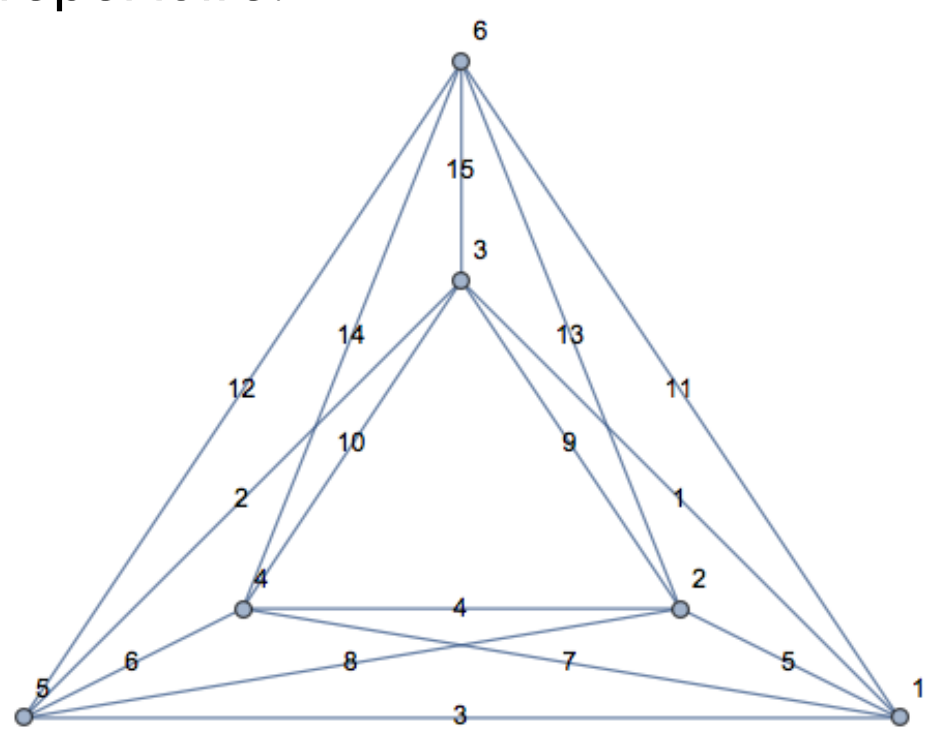

**Figure 12 Symmetric electrode pairing**
Lablet electrodes can be arranged in three groups:
A0 A1 A2
DI DO
PWR2
All pair combinations of these electrodes are depicted graphically in the diagram on the left.

Although autonomous coating control over sensor electrodes would be desirable, this poses significant electronic problems associated with sensitivity which currently could not be resolved with the transistor count available on lablets this size. A solution to this general problem for programmable sensors was found and implemented on the CMOS3 dock [22]. The overall usage of on chip memory resources is shown in Table 3.

| | | |
|---|---|---|
| state | 2 | |
| phase | 2 | |
| patterncnt | 3 | |
| repeatcnt | 7 | |
| clkdivider | 12 | for SYSCLK==200 |
| pulsecnt | 4-5 | counts high time of din value |
| commandin | 8 | receive register |
| trigd | 1 | re/set by commands |
| din1 | 1 | |
| din2 | 1 | |
| comd | 1 | received a command |
| dout | 1 | data out stream |
| doutact | 1 | dout tristate control |
| act | 6 | |
| actenab | 5 | |
| prog | 3*18+4 | |

**Table 2 Lablet state flip-flop counts.**

| Name | length | occurence | comment |
|---|---|---|---|
| REP | 2 | *1 | repeat: 1,4,16,64 |
| DAT | 1 | *1 | save sensor data |
| TIM | 1 | *1 | timestep*: 1,4,16,64 (LSB of TI[1:0]) |
| TI | 1 | *3 | (MSB of TI[1:0]) |
| EC | 1 | *3 | group select |
| SC | 2 | *3 | sequence select |
| EP | 3 | *3 | polarity select |
| CA | 3 | *3 | jump |
| PA | 3 | *3 | pattern |
| NE | 1 | *3 | inversion |
| SE | 4 | *3 | condition |
| sum | 58 | | |

**Table 3 Lablet program memory usage (bits)**

The complete table of program variables with bit counts for lablet function is given in Table 2. A graphical representation of the state variables of lablets is shown in **Fig. 12A**.

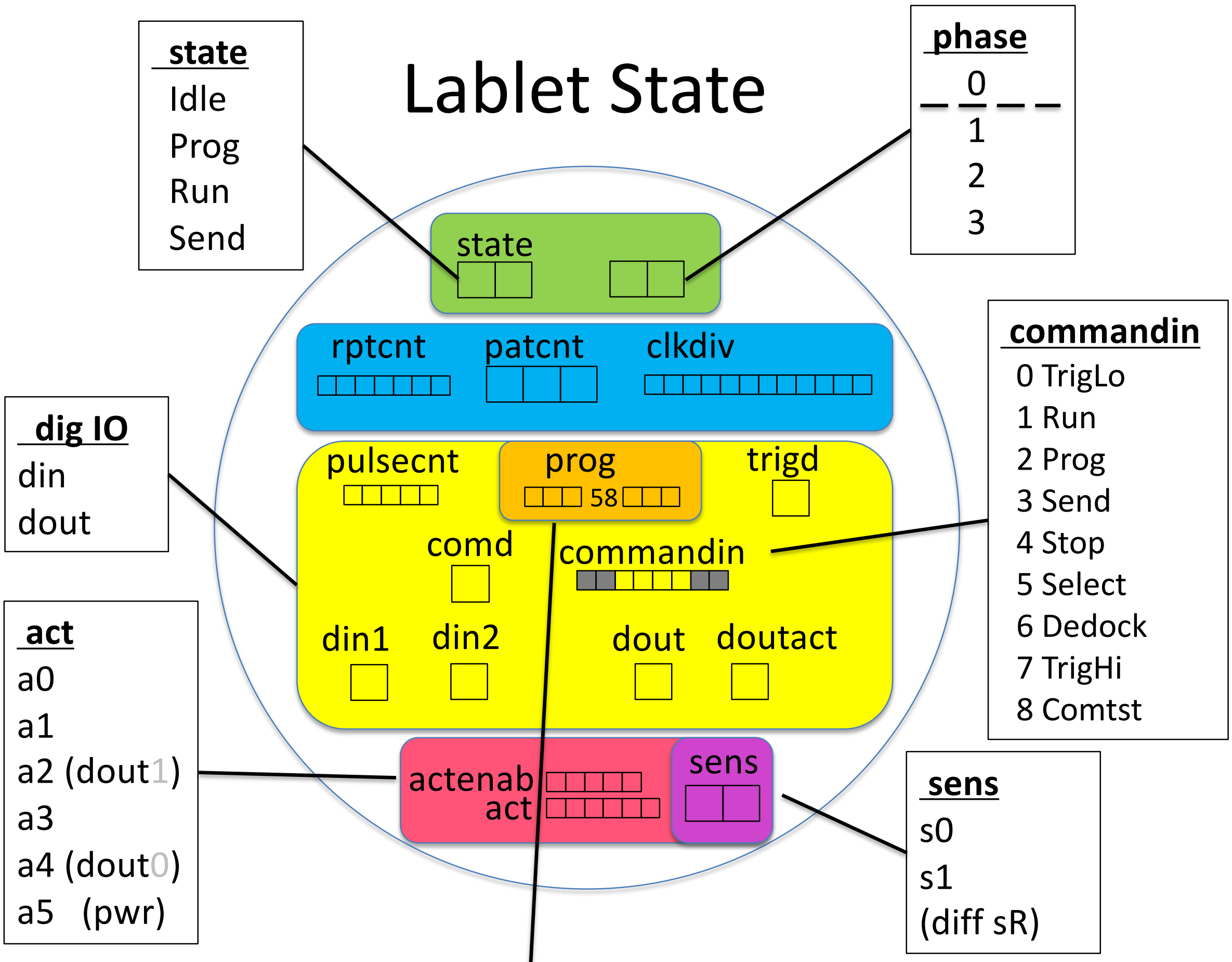

| | | | | | | | | | | | | | | | | |
|---|---|---|---|---|---|---|---|---|---|---|---|---|---|---|---|---|
| ph1: | SC: | 2 | EP: | 3 | TI: | 1 | SE: | 4 | CA: | 3 | PA: | 3 | NE: | 1 | EC: | 1 |
| ph2: | SC: | 2 | EP: | 3 | TI: | 1 | SE: | 4 | CA: | 3 | PA: | 3 | NE: | 1 | EC: | 1 |
| ph3: | SC: | 2 | EP: | 3 | TI: | 1 | SE: | 4 | CA: | 3 | PA: | 3 | NE: | 1 | EC: | 1 |
| all | RPT: | 2 | DAT: | 1 | TIM: | 1 | | | | | | | | | | |

**Fig. 12A Lablet states.** Organized by functionality into four main classes: the main state and phase (green), detailed step counters (blue), command and programming control (yellow/orange), and actuation and sensing (red/purple). The lablet state at any time is recorded in flip-flops (square boxes).

While the above diagrams cover the core functionality of the lablets, they do not deal sufficiently with the intercommunication between lablets and the external world or other lablets. This final set of logic implemented in all lablets is shown in **Fig. 13**. Firstly, lablets are given the ability to initiate programming of other lablets by sending program data. If the program bit TSDAT is set, then the lablets can record sensor values or the time points when threshold events are reached as part of their program data, which can then also be communicated with the outside world (or other lablets) by this mechanism. In addition to the ability to exchange the entire lablet program (58 bits), a shorter exchange of information can be achieved, involving individual lablet "commands" that can be sent or received. The possible commands are shown in green along with their binary communication codes. For parsimony of logic resources, the inner loop pattern generation is sequestered for command IO, and the logic to drive this is also shown in Fig.13.

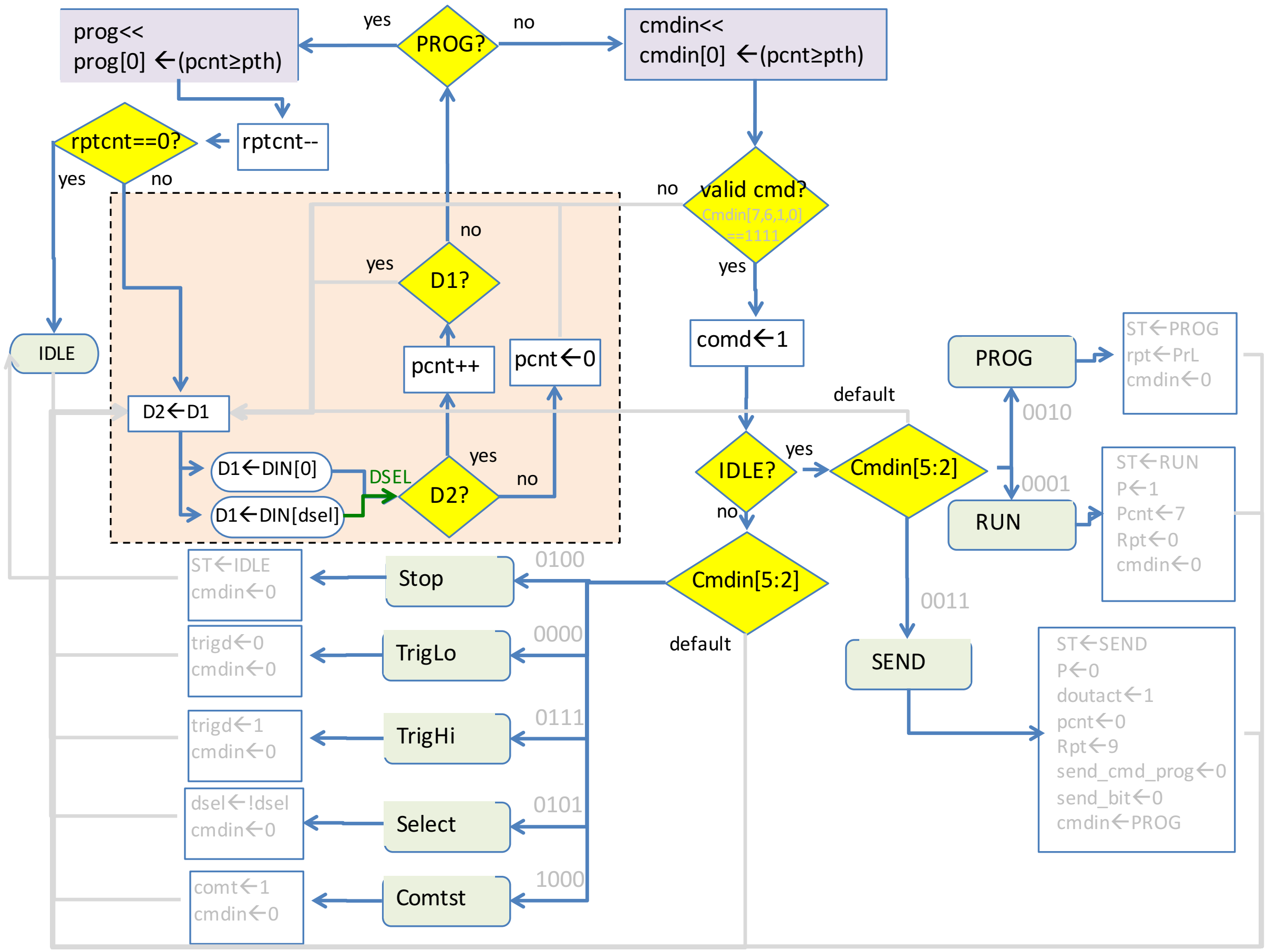

**Figure 13: Lablet communication control logic.** The interaction of lablets with external commands is shown, including trigger, test and stop, in addition to prog, run and send.

The precise Verilog code that is employed to implement this logic in lablets by a mapping to CMOS transistors is presented in the SI (Appendix 3). In cases of any interpretative discrepancies, the Verilog code is the final authority on what was actually implemented in the fabricated lablets. A number of parameters influence lablet logic functionality and communication as shown in table 4.

| Define | id0 | id1 | id2 | id3 | id4 | |
|---|---|---|---|---|---|---|
| SYSCLK200 | + | + | + | + | - | |
| SYSCLK20 | - | - | - | - | + | |
| DEF_ID | 3'B000 | 3'B001 | 3'B010 | 3'B011 | 3'B100 | |
| WITH_ID | + | + | + | + | + | |
| THE_PROG | 'PROG_02 | 'PROG_02 | 'DEF_PROG | 'PROG_03 | 'PROG_04 | |
| LONGCOND | - | - | + | - | + | |
| AUTORUN | + | + | - | + | + | autostart after timeout |
| DDINCHAN | + | + | - | - | - | if din/dout inside channels (Lablet3a) |
| DATA_BIPO | + | + | - | + | + | datasending is bipolar HL and LH on din0/din1 |
| DATA_DCFREE | - | + | - | + | + | then LO pulses are as long as HI pulses on the dataline(s) |
| PULSECNTBITS | 5 | 5 | 5 | 4 | 5 | bits for pulse length counter |
| PULSETHRESHOLD | 4 | 9 | 9 | 3 | 3 | number of clock to determine btw. Hi and Lo |
| PULSELONG | 7 | 16 | 16 | 5 | 5 | no of clks for Hi pulse |
| PULSESHORT | 2 | 2 | 3 | 2 | 2 | no of clks for Lo pulse |
| PULSEPAUSE | 3 | 7 | 7 | 1 | 1 | no of clks btw. pulses if not DCFREE |

**Table 4: Lablet logic variant parameters in CMOS fab.** The table shows the values of variable parameters in the digital logic for lablet types: ID 0-4. For example, the slow lablet clock runs at either 20 or 200 Hz (nominal value).

Optical bar codes were used to encode lablet variants. The linear code employed in CMOS2 at 2µm scale proved challenging for large area imaging microscopes and was replaced by a lower resolution (5µm) 2D code. A catalogue of lablet designs contains further details of lablet designs and the different bar-coded variants. Optical bar codes will be useful for experimentalists not only in confirming lablet identity but also to enable applications such as lablet sorting of cargo.

Finally, the mapping of lablet variants to silicon wafers is mastered using a programmable approach, since hundreds of different variants need to be coordinated. The logic is mapped to a cell library for the 180nm technology and placed in a modularly defined region on the lablet, which juxtaposes and matches up with the analog circuitry (section 2). The complete lablet circuitry for digital logic is placed in a catalogue of combinatorial design elements and then assembled via a scripted language program to produce all the mask layers for a single reticule on the wafer. This reticule is stepped at fabrication time over the wafer. The reticule size (ca. 11x14 mm) is chosen so that four reticules fit into a single illumination area, so that compound masks serving 4 layers at a time (via 4 exposures with offset) can be used for the circuit. This greatly reduces the cost of fabrication.

## 5. Fabrication of CMOS3 lablet variants

Electronic fabrication was performed in 180nm technology on 200mm silicon wafers based on the custom designs presented above, using Europractice with (IMEC, Belgium) and the foundry (TSMC, Taiwan) in an engineering run (12 wafers), as in the CMOS2 fabrication. Memcap capacitors were not used, to avoid topographic variations and additional filler structures in top metal, required to meet fabrication constraints, were removed to the base of walls and the sacrificial zone between lablets. This allowed the CMOS fabrication to restrict topographic variations to those essential for microelectrodes or to features commensurate with the desired topography of processed lablets. This is why the unprocessed lablets already have the topographical appearance of processed lablets (apart from the sealing laminate).

For the fabrication of the CMOS3 lablets, a modification of the main fabrication scheme employed in CMOS2 lablet fabrication [10] is employed, as shown in **Fig. 14**. It is based on the differential galvanic coating device developed in [10] to overcome problems in making global electrical connections between individual reticules (illumination zones) on the silicon wafer because of steep-walled topographic features in the inter-reticule zones (TSMC test structures). After the initial ENEPIG coating (step 1), and optional photolack protection of actuator electrodes that should remain with an upper Au coat (step 2, mask 1), the central differential galvanic process is carried out on a reticule by reticule process. Following galvanics, the protective photoresist is removed and then SU8 walls are assembled (step 3, mask 2). Next the electrolyte is laminated into the supercaps and the channels are lidded with the laminate, selectively photostructured where required (step 4, mask 3), before the wafers are thinned and singulated.

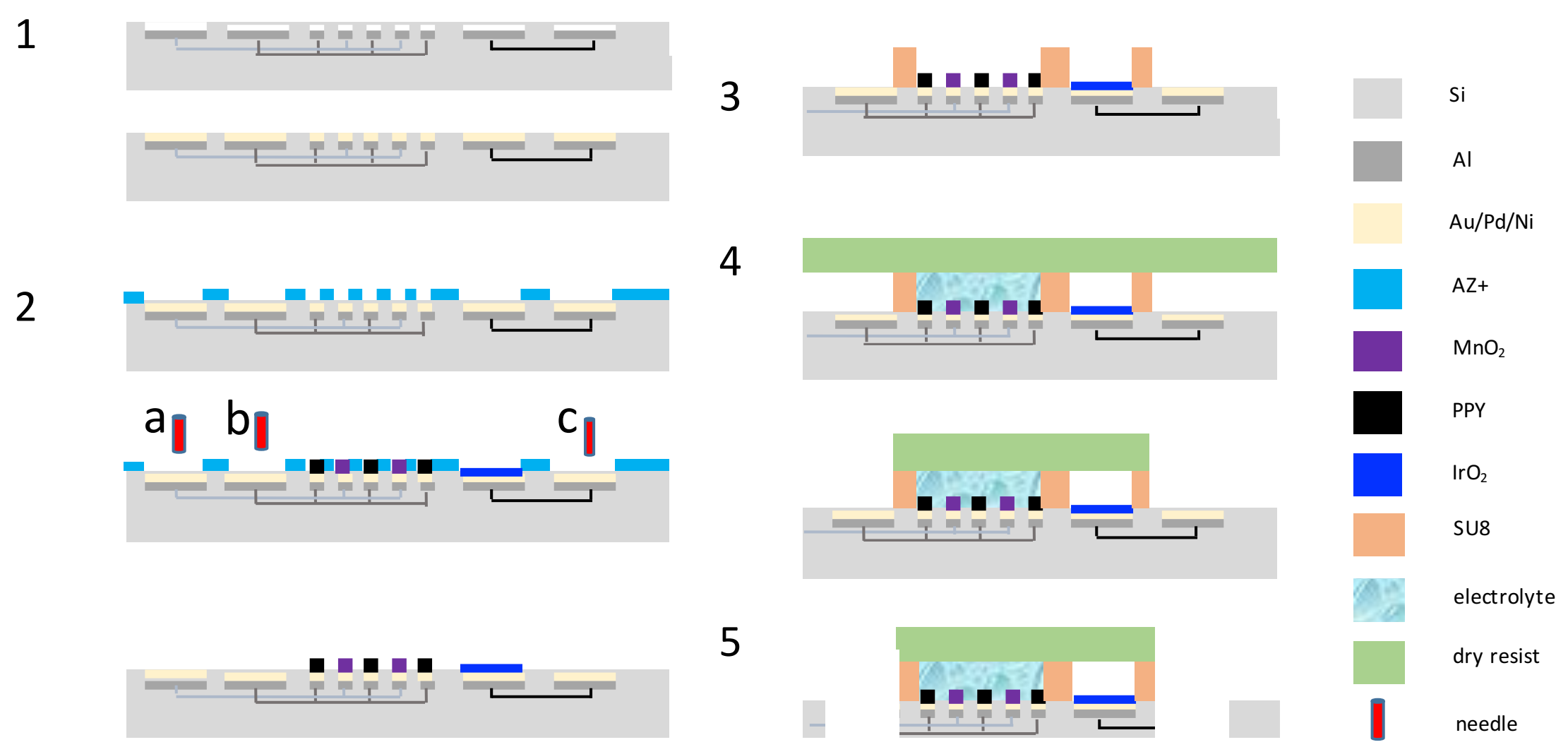

**Fig. 14. Modified post-processing scheme for lablets.** (1) electroless metal fill deposition to gold coated microelectrodes, ENEPIG, (2) differential galvanic coating, optionally protected by positive resist (3) SU8 walls (4) electrolyte encapsulation and photostructured lamination (5) wafer thinning and sawing for singulation of lablets. Step 1 is not required with individual reticule galvanics: this is more controlled but less efficient (see text).

The results of the fabrication process are documented in the following sections, describing the fabrication briefly, highlighting differences with the CMOS2 fabrication [10].

## 1. ENEPIG coating

The ENEPIG procedure (courtesy of Atotech GmbH, Berlin) was performed with thicknesses of 450/250/70 nm for Ni/Pd/Au respectively. The ENEPIG coating (see **Fig.15**) shows a fine-grained gold top layer that covers the pads of the lablets without alignment errors, because of the maskless autocatalytic procedure.

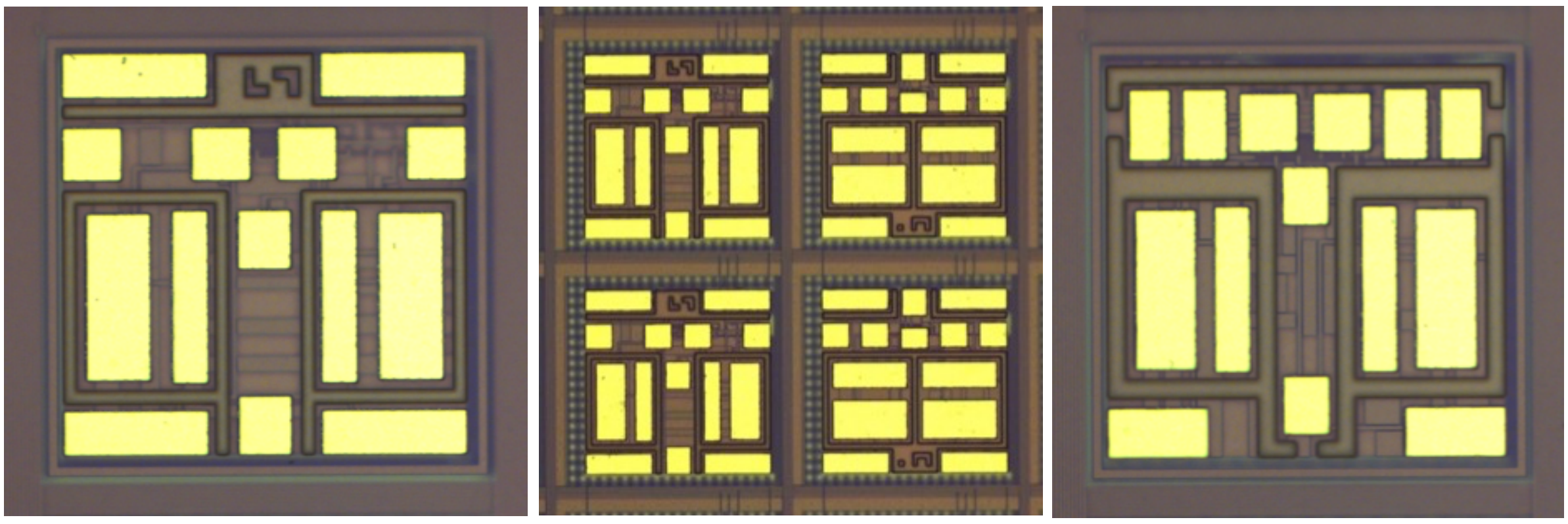

**Fig. 15 ENEPIG coating of the three different CMOS lablet types (1a,2a,3a.** Lablet dimensions 140x140 µm. The center image contains 4 lablets of type 1a and 2a, which are labelled with an optical bar code, readable with ≤4 µm resolution (to mark different variants). The wall structures are already visible, being raised as a wall foundation by top metal structures (under the isolation layer). Also visible in the centre image are the bus connections used for differential galvanics, connecting corresponding electrodes for all lablets on a reticule.

### Initial singulation (short process)

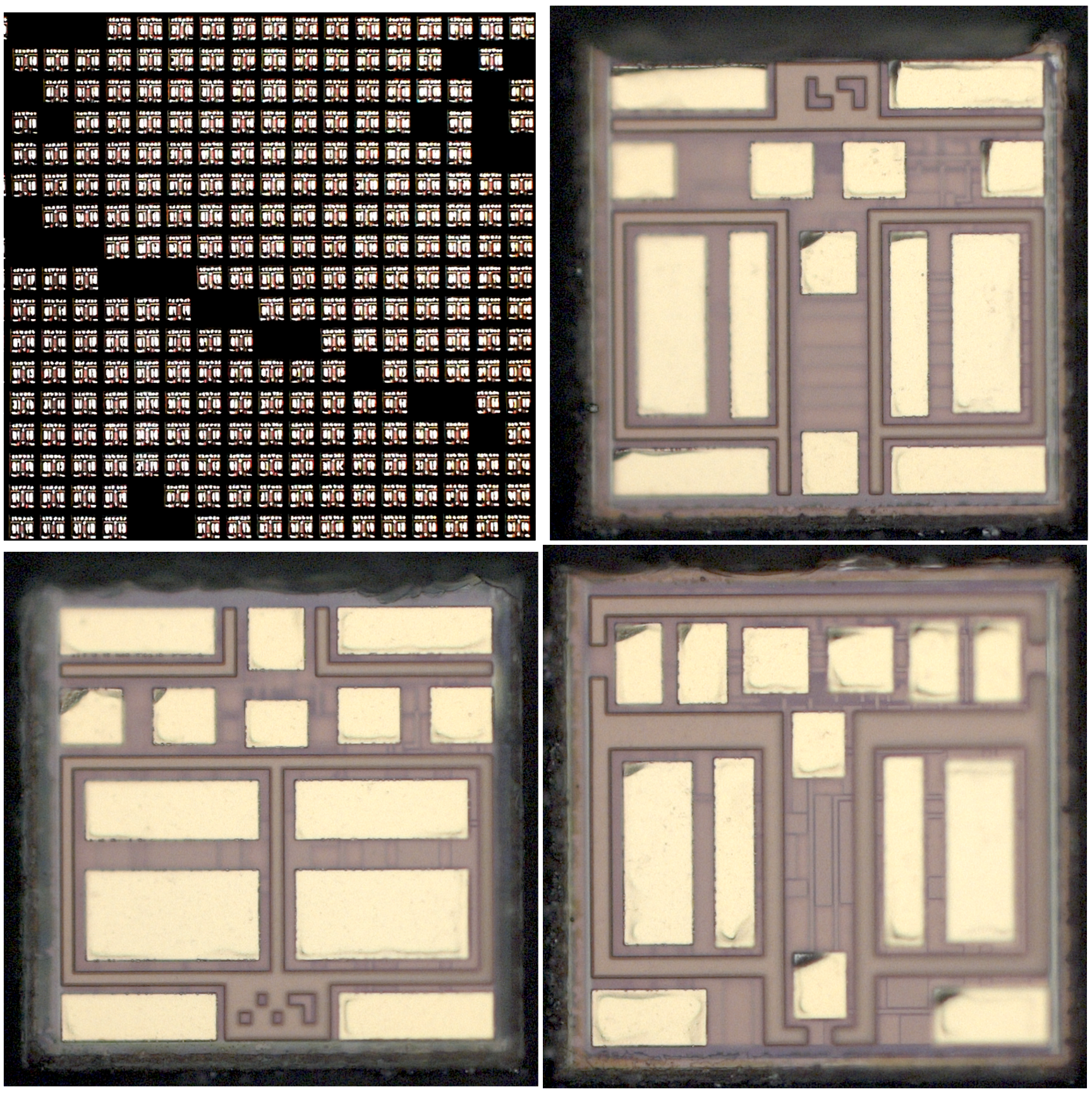

**Fig. 16. Singulation of ENEPIG lablets TL:** Diced lablets on the transfer tape, showing loss of some lablets during tape transfer after grinding. TR,BL,BR Three different lablet types (1a,2a,3a) after singulation. The first singulation run was performed with a combination of laser cutting through the upper layers of metal and then fine mechanical sawing of deeper grooves before grinding from the back side. This still resulted in considerable chipping at the edges of lablets. In addition, some detachment of edges of the ENEPIG coat are discernible.

After this coating, a first wafer is sent to DBG (dice before grind) singulation (Disco GmbH, Germany), to gain both singulated lablets and lablet arrays. The process resulted in a reasonable yield of intact lablets, with some losses and chipping of lablets, and a certain amount of detachment at the corners of the ENEPIG coat (see **Fig. 16**). The large amounts of metal in the separation lanes prevented a direct sawing approach (blade deviation and breakage), and so initial laser cutting was employed (E. Eurich, Disco).

## 2. Differential galvanics

The photo lack protection of actuator electrodes is carried out with removable positive photoresist (ca. 4 µm, AZ+, step 2, mask 1). The mask provides barriers to galvanic growth also laterally between the close lying electrodes (supercap and sensors) allowing thicker galvanic coatings to be applied. Briefly according to the method developed in [10,14], up to 9 reticules at a time are contacted with a custom parallel electrical probe adaptor unit mounted on an xyz-stage, allowing the different types of coatings to be applied successively through translation of the device, first to different 3x3 reticule zones of the wafer and then, after changing the electrolyte solution, to the next coating family of electrodes.

An example of the differential galvanic coating for supercaps is shown in **Fig. 17**. For more details of the apparatus and procedure see also [14].

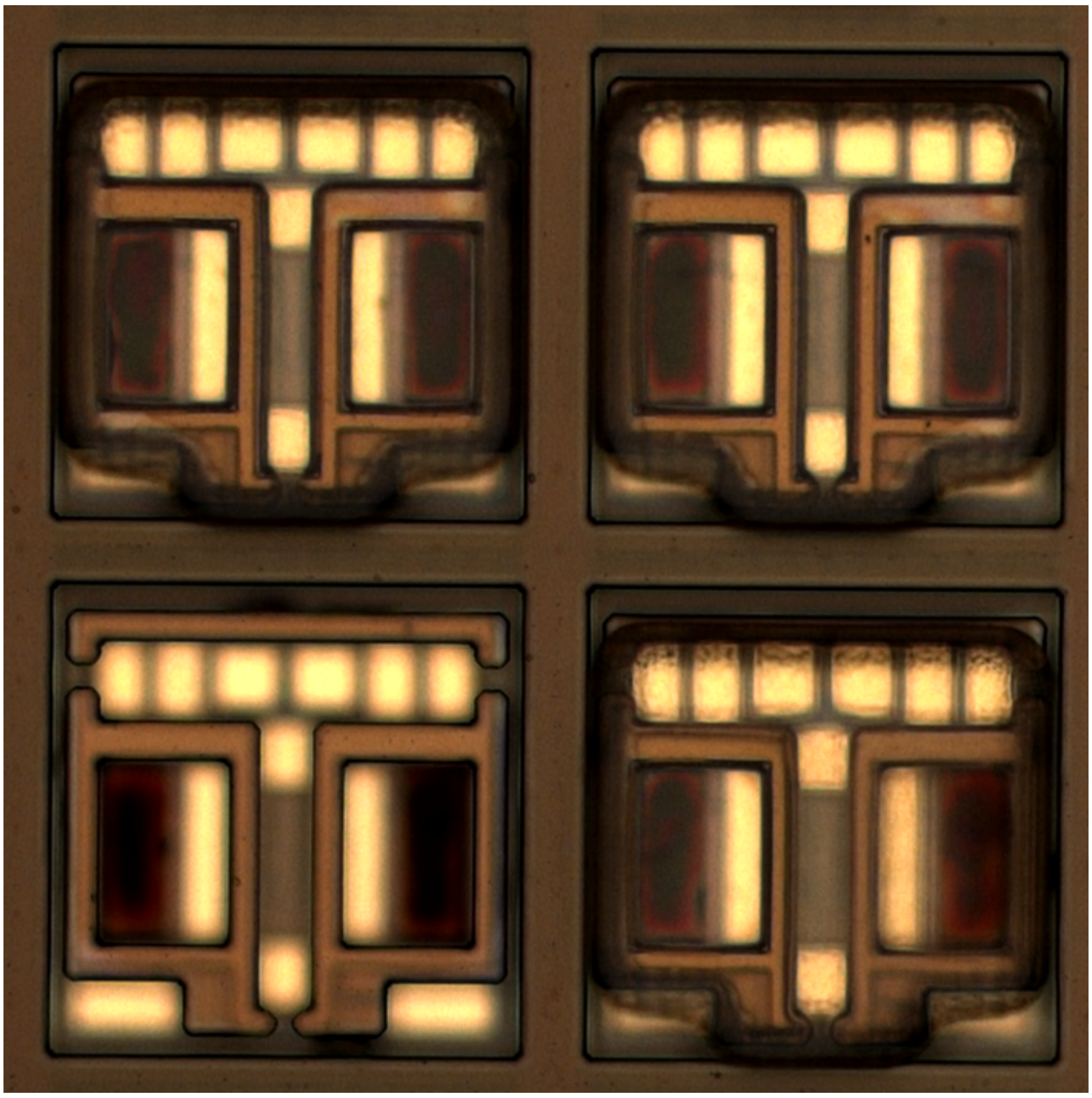

**Fig. 17: Images of specifically galvanically coated lablets.** The two supercaps on each of the four lablets shown here are coated in parallel asymmetrically at the level of reticules in parallel. Here only on the larger electrodes (darker colour) are coated, leaving the smaller rectangular electrodes (adjacent in gold) bare**.** Both symmetric $MnO_2$ and asymmetric $MnO_2$/PPY coating pairs were employed in successive galvanic coating steps for the final lablets. Three of the lablets have intact laminate cover layers.

## 3. SU8 wall structures

The SU8 walls were fabricated with base 9 µm and height 7µm (later to be extended to 12-15 µm) using conventional spin coating photolithography (step 3, mask 2, Temicon GmbH, Dortmund, Germany) as in [10]. The mask for step 3 was augmented with additional wall structures in the sacrificial areas between lablets to increase the adhesion of the laminate, which has to sustain multiple carrier tape removal processes, once for the laminate-liner and once during singulation. The photolaminates were developed according to a low temperature variant of the manufacturer's protocol as in [10]. Example images of the SU8 wall structures are shown in **Fig. 18**.

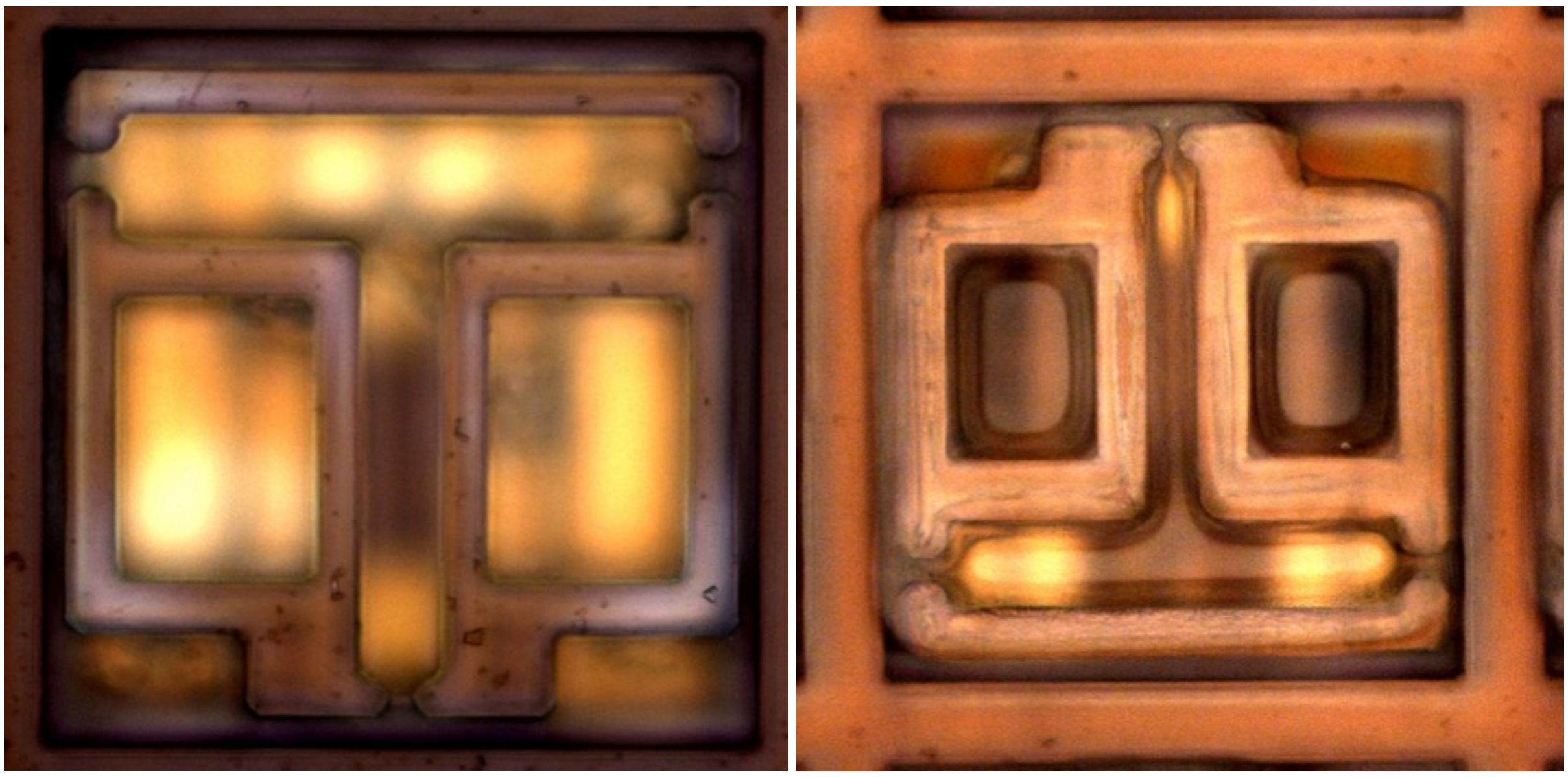

**Fig. 18 SU8 wall structures for lablets. Left:** Using a 50x objective, the SU8 wall structures with laminate lablet were photographed. The image shows the important role of the sacrificial square mesh walls to increase the surface area for wall adhesion during laminate peel off. **Right:** SU8 wall structures in focus underneath laminate. The laminate can also just be discerned (out of focus) as well as the electrolyte partially filling the supercap.

## 4. Photolamination and electrolyte encapsulation

In step 4, electrolyte solution is applied in a line of drops on the wafer and photolamination (here with Nagase DF1010 or the more hydrophobic DF3014) proceeds as described in [10] at 80°C to seal in the electrolyte to the supercap structures. All other structures are open to the outside so that excess electrolyte can be removed by drying and rinsing. As electrolyte, two different salt solutions were used at concentrations of 1M $Na_2SO_4$ and $KNO_3$, together with the additives optimized in [10]: 50% glycerine (hygroscopic) and 0.025% triton-X (surfactant).

Whereas $Na_2SO_4$ is a good electrolyte for the more readily produced symmetric $MnO_2$ coated supercaps, these have a too limited voltage window (1V), and instead an asymmetric combination of $MnO_2$ and PPY (polypyrrole) is used as described in [15]. The asymmetric supercap performance is significantly better in $KNO_3$.

The fully processed lablets, including wet lamination on the 8" wafer to enclose the electrolyte, are shown in **Fig. 19**. Note that electrolyte was applied only to a portion of the array of reticules, to allow for dry reference structures for comparison. Details of the so processed lablets can be discerned only in magnification, as shown in **Fig. 20**.

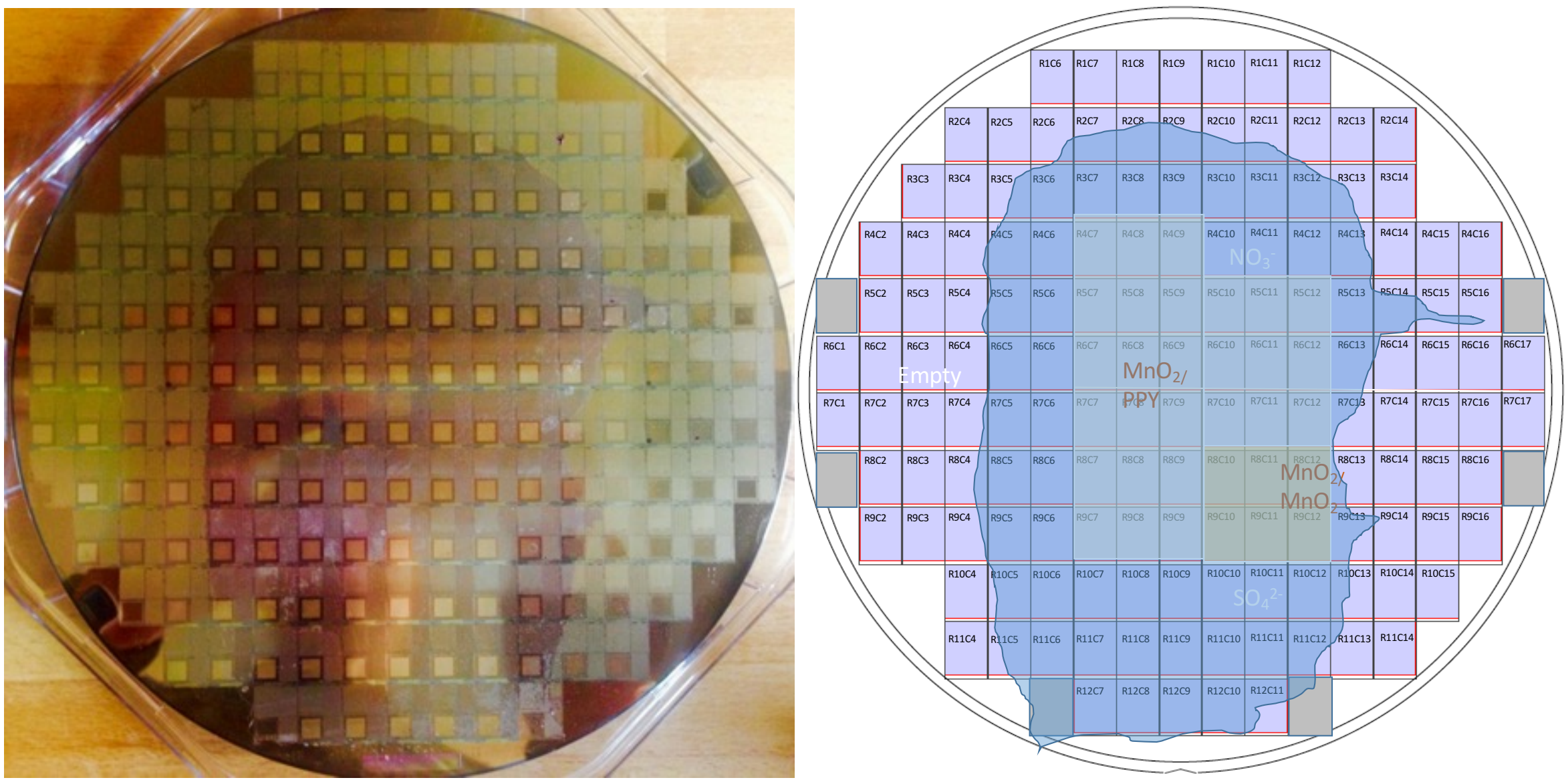

**Figure 19: Fully post-processed 200mm CMOS3 wafer. Left:** 160 reticules are shown, each including both one dock (dark rimmed squares) and thousands of lablets (whitish from photo-structured laminate). The wafer has been through the full processing steps (see text) and (two different) electrolyte solutions have been encapsulated with the differentially coated supercaps, showing dark in the central region. **Right:** Map of electrolyte and coating distribution on the wafer. The wafer is cut either for lablets or for docking chips and lablet arrays. Gray peripheral reticules are not completely within processed area of wafer. Lamination proceeded from left from a vertical row of electrolyte droplets (1M $KNO_3$ or $Na_2SO_4$ aqueous solutions in 50% glycerine), allowing filled and unfilled lablets structures to be compared. Both symmetric and asymmetric coated supercaps were produced.

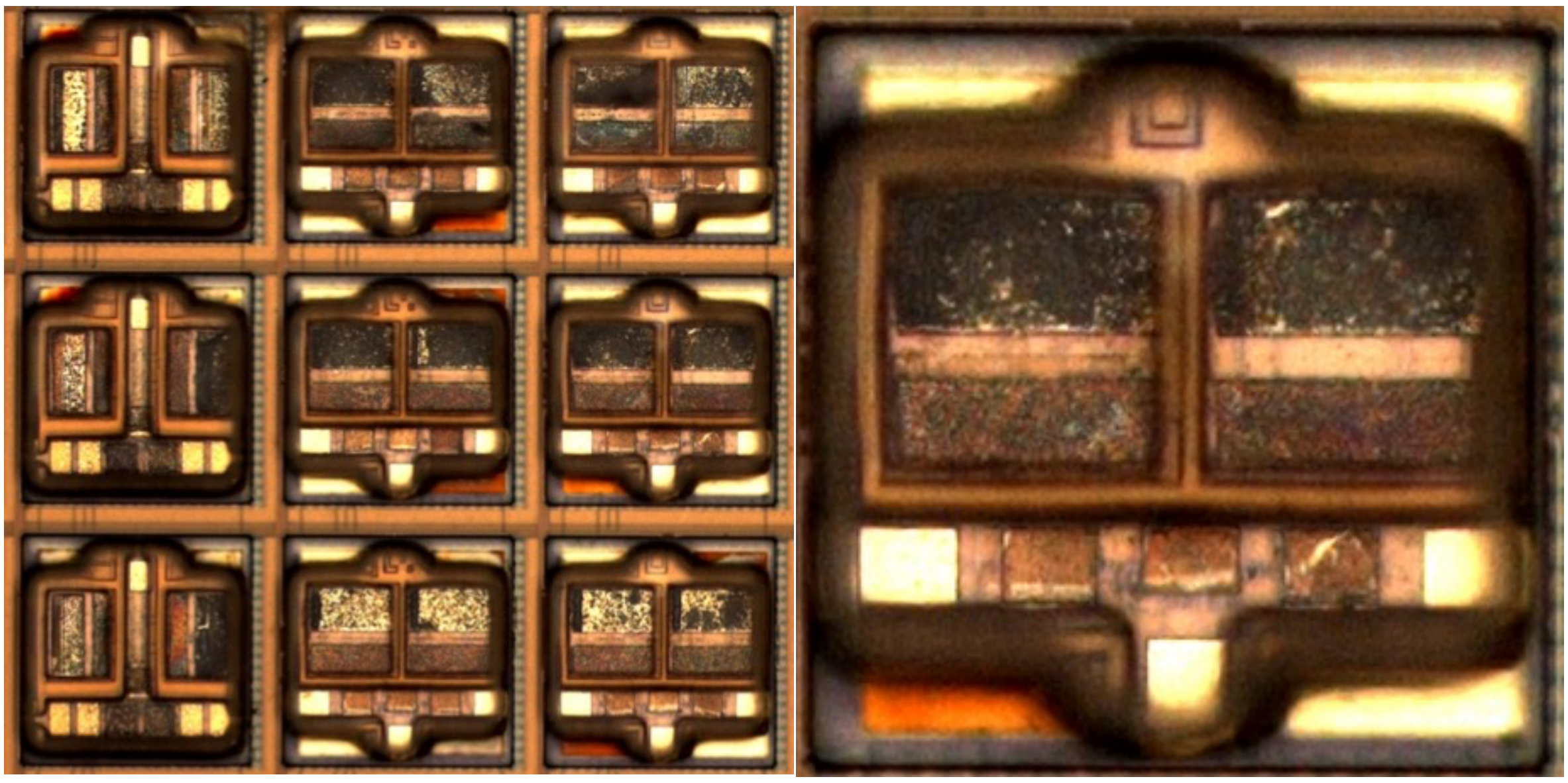

**Fig. 20. Laminated and coated lablets. Left:** Six lablets of type 2a and three of type 3a are shown following development of photostructured laminate (step 4). The asymmetrically coated supercaps and uncoated actors are clearly discernible beneath the transparent laminate (DF1010). Note also the sacrificial square mesh of SU8 walls between lablets, designed to minimize stress on lablet walls. **Right:** magnification of one of the lablets. Lablet 140x140 µm.

## 5. Lablet singulation

Singulation of the CMOS3 lablets was carried out by (Disco GmbH, Germany) using a laser cutting process with ca. 20µm wide grooves together with a back grinding process, employing a low-binding carrier tape to minimize stress, to thin the wafers to 30 µm Si. While this procedure worked well for lablets up to the SU8 wall structure, it had a very low yield of intactly laminated lablets, as can be seen in **Fig. 21**.

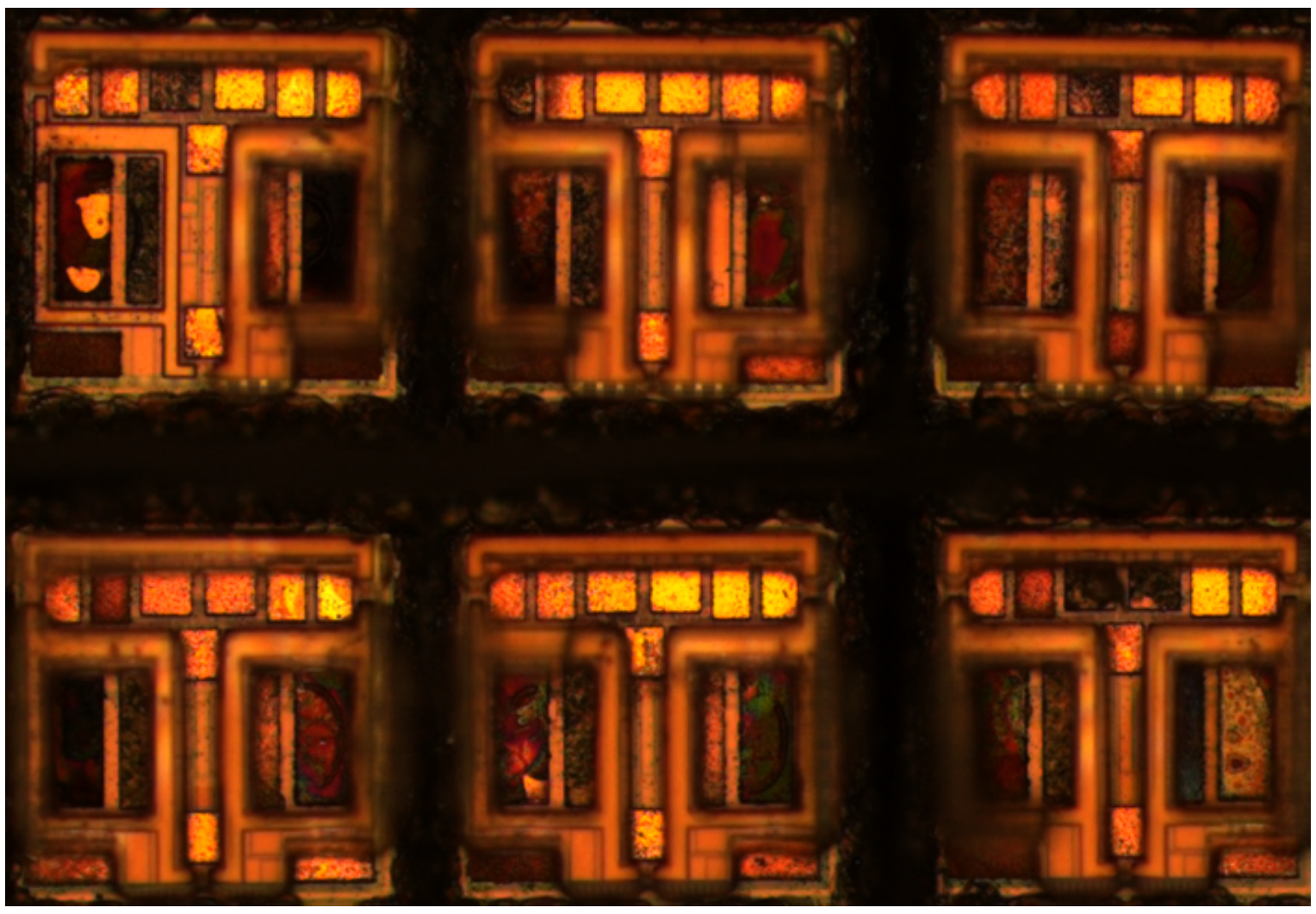

**Fig 21 Singulated fully processed lablets.** The singulation process damaged many of the lablets, here leaving the SU8 walls mostly intact but removing the laminated lids from the lablets.

The lablets were also processed to lablet arrays for systematic investigation with some externally wired contacts. There, the lablets remained intact even after dicing, including the encapsulated electrolyte as shown in **Fig. 22**.

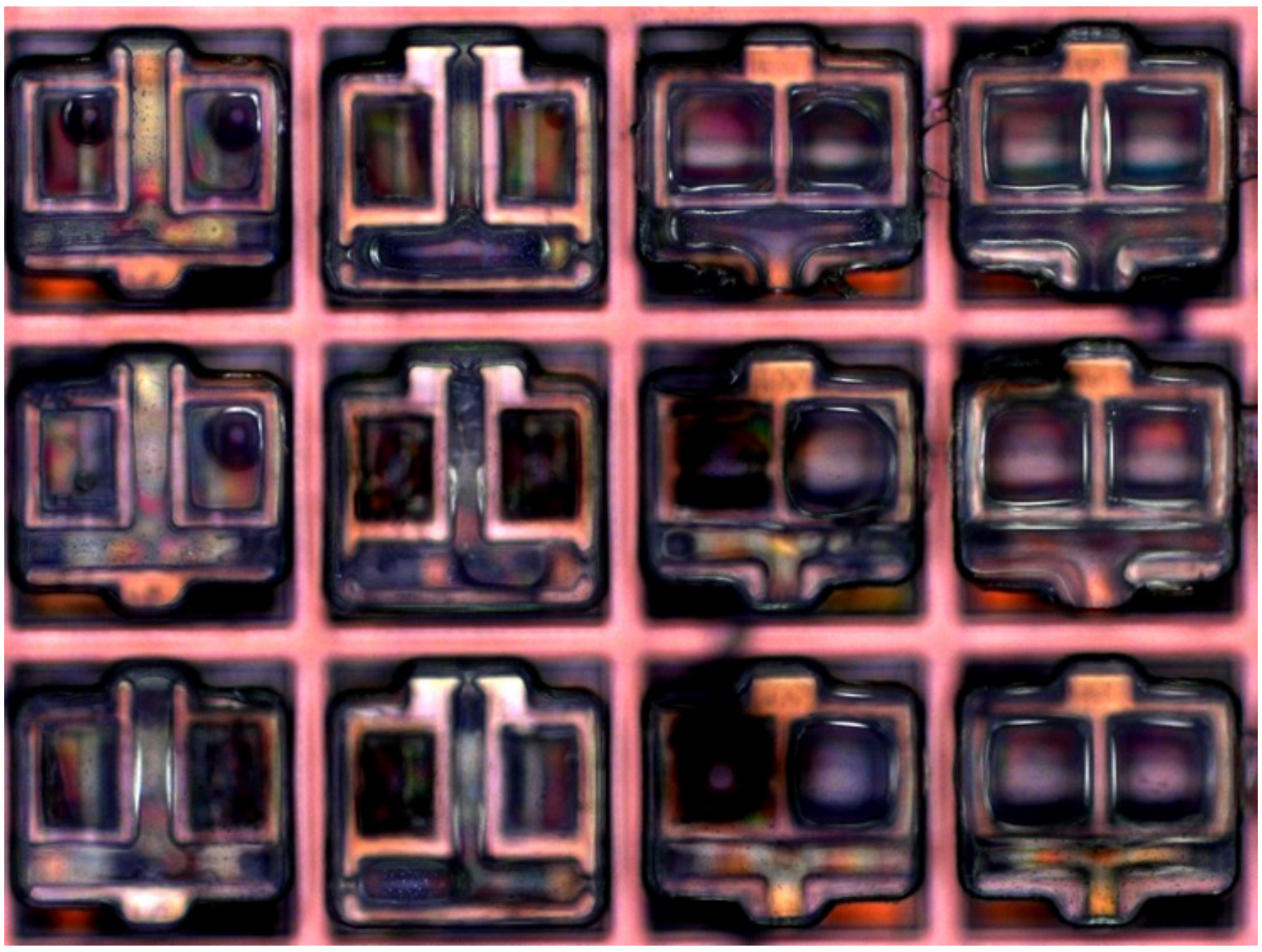

**Fig. 22. Lablets, complete with encapsulated electrolyte supercaps on lablet array.** The lablets in the array show the encapsulated electrolyte, visible in places because of partial filling (bubbles).

# 6. Autonomously clocked operation of individual lablets tested using electrochemiluminescence

Like their CMOS2 predecessors, the electronic functionality of the CMOS3 lablets were tested both in singulated form and in parallel on lablet arrays: where they share a common power supply. For the layout of variants in CMOS3 lablet arrays see SI Appendix 4. Since the lablets each have their own ultra-low power slow clock, with slightly different periods, after a synchronous power on, one expects otherwise identical lablets to exhibit initially similar actuation patterns, which slowly diverge in time because of phase differences. Because of the large number of different lablets, a method to test them all in parallel was needed. As in previous work [9,14] the electrochemiluminescence reaction of the $Ru(bpy)_3^{2+}$ system, with a tri-n-propylamine (TPA) sensitizer [23]. At voltages above 0.8 V, the redox process creates an excited $Ru^{2+}$ ion, which then emits (wavelength 610nm) according to a complex mechanism. This allows the electrochemical functionality of voltage switched actuator electrodes to be verified in parallel while submersed in aqueous solution, optically (**Fig. 23**).

The supporting information, SI Appendix 5 contains a video sequence example of such recorded spontaneous switching of lablet electrodes, running their individual default programs (see also Appendix 3). As in earlier work with CMOS2 lablet arrays, the patterns of switching can be checked against simulated temporal switching patterns for the different lablet variants (including different slow clock rates). These observations confirm the general correctness of the lablet logic (although sensor functionalities have not yet been checked). Further verifications of programmed functionality on singulated CMOS3 lablets have also confirmed this and will be published separately.

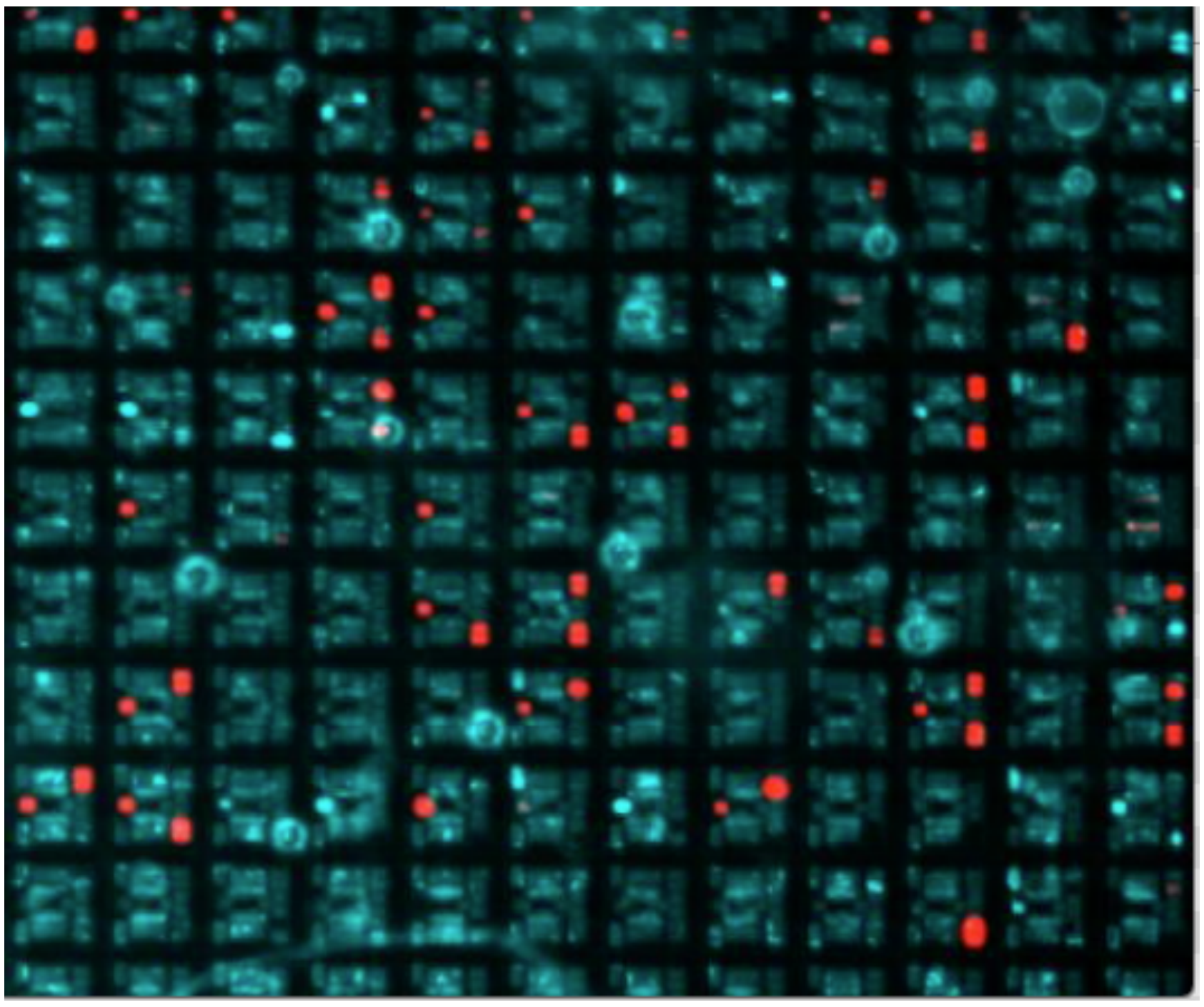

**Fig. 23 Electrochemiluminescence of CMOS3 lablets.** The lablets shown are a small portion of one of the ca. 1000 lablet arrays. The snapshots shows one time instant. A video of the programmed switching is presented in SI Appendix 5.

## 7. Discussion and Conclusions

This paper has reported the design and fabrication of an optimized family of electronic lablets (CMOS3) with full post-processing functional integration able in principle to support a rich spectrum of lablet functionalities (see Section 2, 7). The digital architecture provides a compact solution to maintaining rich programmability, and the ability to proliferate and evolve lablet programs by communicating them between lablets, while restricting resources to a range of actuator and sensing timing patterns that are likely to be useful in interacting with chemistry. Only by this means was it possible for example to condense the complex functionality required for a lablet life cycle, to allow it to be programmed with a 60-bit program memory without restricting the flexibility of the lablet for other tasks.

Although the lablets are currently restricted to binary threshold sensing, as discussed, the combination of sensors with timing information and/or modulation via on lablet actuators can deliver a wealth of analog information. The exploration of the functional integration of autonomous sensing and actuation is still in its infancy and can only really take off when reliable autonomous power on the lablets is achieved. Further work will deal with the higher order functionalities built into the lablets, as listed in Section 2.

The encapsulation of electrolyte is despite multiple optimisations not reliable and further work on this is required: optical proof of encapsulation was obtained for a few lablets only. Thus, despite successful integration of galvanic coating, full supercap functionality has only been achieved so far in open electrolyte lablets with the same dimensions [15].

A non-trivial example of lablet program functionality, coupling sensors to locomotion in chemotactic behaviour has been published [18]. Many further functional programs, including self-replication programs from one lablet to the next, involving both copying of electronic program information and customization of the bare lablet offspring electrodes with programmed coatings are possible.

*Acknowledgements:*
This research is supported by the European Commission, EU FET Open MICREAgents Project # 318671. Chemical preparations and wafer handling were assisted by J. Bagheri-Maurer. We acknowledge the accommodating support of P. Malisse and J. Verherstraeten at IMEC, especially in Europractice design rule checking close to Christmas. The authors wish to thank M. Richter and Atotech GmbH for complimentary electroless coating of wafers; S. Kaiser, M. Fleger and A. Lamhardt for extensive customer support during wafer processing at Temicon and for hosting our work on lamination in their clean rooms; and E. Eurich at Disco for his willingness to explore new options in wafer dicing for our wafers. Special thanks also go to N.N, N. Plumeré and W. Schumann at Electrochemistry, RUB for advice with pipette pulling for reference electrode construction. Our thanks are also extended to K. Kallis TU Dortmund and A. Ludwig at RUB for their willingness to plan alternative fabrication routes with us. Finally, our thanks go to all the partners in the MICREAgents project for their willingness to co-engage with lablets and for inspiration in how chemistry may be linked with them. This manuscript was submitted in 2016 to the EU for external review as part of the final review of MICREAgents project, and has been reformatted here for ArXiv publication with minor updates to the references for those articles then in preparation and current first author affiliation.

**Supporting Information: list of appendices**

Appendix 1 Instruction tables for three minimized lablet microcontrollers
Appendix 2 Verilog code for digital lablet logic generation: with 3 attached files
Appendix 3 Simulation of lablets via testbench
Appendix 4 Lablet array variant keys
Appendix 5 Video sequence of electrochemiluminescence of lablets in array.

# Appendices (SI)

## Appendix 1 Instruction tables for three minimized lablet microcontrollers

See SI for verilog implementation in assembler and subsequent mapping to CMOS.

1. Instruction table for lablet-optimized 4-bit microcontroller with Harvard architecture.

| instruction | op | reg | operation |
|---|---|---|---|
| out_ea | 0 | | accu -> 4 bit actor |
| sensein | 1 | | 4 bit sensor -> accu |
| din | 2 | | 4bit commdata -> accu |
| wait | 3 | | wait(accu) |
| xchab | 4 | | accu <-> breg |
| test_skip | 5 | | jump +1 if Z |
| and | 6 | | accu ^ breg -> accu, Z |
| or | 7 | | accu v breg -> accu, Z |
| xor | 8 | | accu xr breg -> accu, Z |
| | 9 | | |
| st | a | | accu -> mem[breg] |
| ldi | b | xxxx | xxxx -> accu |
| jmp | c-f | xxxx | Jump xxxxxx |

pc xxxxxx
accu xxxx
breg xxxx
zflag x

Microcontroller resources and connections to actors and sensors

16 Regs 4bit
64x8bit memory
PC 6bit
Alu Accu
Ex = 0..14 Z actors
Sx sensors
Jump A/B or C/D

## 2. Instruction table for a potential future 4-bit custom stack microcontroller.*

| | | | N1 | N2 | |
|---|---|---|---|---|---|
| 0 | literal from mem | lit | >N1 | | Push literal value from memory onto data stack |
| 1 | pop | pop | >N2 | >N3 | Pop from data stack, (N2->N1, N3->N2, N4->N3) |
| 2 | store | st | N1> | ADR | Store N1 at address ADR |
| 3 | swap | xch | >N2> | >N1> | Swap N1 and N2 |
| 4 | Shift left | shift | >N1> | | Cyclic shift by one bit left |
| 5 | bitwise and | and | N1> | N2> | Place bitwise and result in N1 |
| 6 | bitwise xor | xor | N1> | N2> | Result place in N1 |
| 7 | greater than | gt | N1> | N2> | 0 if N1>N2, 1 otherwise |
| 8 | read communication | read | >N1 | | Reads in last 4 bits from communication |
| 9 | exit subroutine | end | | | Exits from subroutine, return to prev. PC from RS. |
| 10 | skip if 0 | skipif | N1> | | Skips next operation if N1=0 |
| 11 | loop exponential | loopx | N1> | | Repeat next cmd $2^{N1}$ times. Also used as wait. |
| 12 | set active block | blk | N1> | | Choose 1 of 4 electrode blocks for masking |
| 13 | set two group active masks | mask | N1> | N2> | Set electrode masks for A and B groups in current block. For sensors : sig/ref. |
| 14 | out to active electrode groups | outel | N1> | N2> | 2*2 bits specify A and B group output values (0,1,Z) in active block. Option: N2 time. |
| 15 | read of sensor electrodes | inel | ->N1 | | Read 2*2bit from sensor block, signal and reference. |

### Processor

The current implementation is optimized for extremely low total memory and logic resources on a tiny lablet chip (100*100 µm). The aim was a complete implementation with less than 256 flip-flops including processor and periphery.
8*4 bit data stack
4*4 bit return stack
32*6 bit program memory

### Periphery

Up to 16 electrodes in 4 blocks of 4 electrodes.
*System* : VCC, GND, DIN, DOUT
*Pattern*: P0,P1,P2,P3
*Actors*: A0,A1,A2,A3
*Sensors*: S0,S1,S2,S3

Our periphery state requires 46 bits.
Each electrode has 2 bit state : high/low and output enable. Up to 32 bits.
1 2-bit active block register
2 dedicated registers : output masks A & B . 2 * 4=8 bit.
1 dedicated input register : 2*2=4 bits

*Adapted from a conventional stack microcontroller instruction set [11].

## 3. Instruction table for a 4-bit custom evolvable microcontroller.

The third microcontroller footprinted was a microcontroller design employed as a reference point for the development of an evolvable microcontroller [12].

| instruction | operation | asm | alu | 2reg | transfer | | | flag/PC | | | | | | | |
|---|---|---|---|---|---|---|---|---|---|---|---|---|---|---|---|
| 0 | act_Eout {bit} | st acc,regE | | 1 | acc | -> | reg | | | -> | ld | acc,reg | st | ac | c,Ereg |
| 1 | set_ctr {bit} | st acc,regC | | 2 | acc | -> | reg | | | -> | ld | acc,reg | st | ac | c,ctrX |
| 2 | and {reg. nr} | and acc,reg | 0 | | acc&reg | -> | acc | | | | | | | | |
| 3 | or {reg. nr} | or acc,reg | 1 | | acc\|reg | -> | acc | | | | | | | | |
| 4 | not {reg. nr} | not acc,reg | 2 | | !acc | -> | acc | and 2er complement | | | | xor 0xF | (add 1) | | |
| 5 | xor {reg. nr} | xor acc,reg | 3 | | acc^reg | -> | acc | | | | | | | | |
| 6 | shiftr {nr. bits} | shf acc | 4 | | acc>>1 | -> | acc | | | | | | | | |
| 7 | gotol {reg. nr} | goto reg,nr | | | reg | -> | PC | if 1 | | | | | | | |
| 8 | ifz {accu} | gotoz reg,nr | | | reg | -> | PC | if Z | | | | | | | |
| 9 | ifnz {accu} | gotonz reg,nr | | | reg | -> | PC | if NZ | | | | | | | |
| a | st {reg. nr} | st acc, reg | | 0 | ACC | -> | reg | | | | | | | | |
| b | ld {reg. nr} | ld acc, reg | 5 | | reg | -> | acc | | | | | | | | |
| c | wait {ctr. nr} | | | | | | | if Ctr! | 0 | | | | | | |
| d | set_accu {val} | ldi acc, 4bit | 6 | | immediate | -> | acc | | | | | | | | |
| e | sub {reg. nr} | sub acc, reg | 7 | | acc-reg | -> | acc | leave out | | (not, add 1, add reg) | | | | | |
| f | add {reg. nr} | add acc, reg | 8 | | acc+reg | -> | acc | | | | | | | | |

## Appendix 2 Verilog code for digital lablet logic generation

See three separate Verilog files on github repository named lablet_fsm3 https://github.com/js1200/lablet_fsm3.git:

| | |
|---|---|
| lablet_fsm3.v | main Verilog file for finite state machine CMOS3 |
| lablet_fsm3_defs.v | include file with definitions and variants id0, id1 … id4 |
| lablet_fsm3_tb.v | testbench file |

| Define | id0 | id1 | id2 | id3 | id4 | |
|---|---|---|---|---|---|---|
| SYSCLK200 | + | + | + | + | - | |
| SYSCLK20 | - | - | - | - | + | |
| DEF_ID | 3'B000 | 3'B001 | 3'B010 | 3'B011 | 3'B100 | |
| WITH_ID | + | + | + | + | + | |
| THE_PROG | 'PROG_02 | 'PROG_02 | 'DEF_PROG | 'PROG_03 | 'PROG_04 | |
| LONGCOND | - | - | + | - | + | |
| AUTORUN | + | + | - | + | + | autostart after timeout |
| DDINCHAN | + | + | - | - | - | if din/dout inside channels (Lablet3a) |
| DATA_BIPO | + | + | - | + | + | datasending is bipolar HL and LH on din0/din1 |
| DATA_DCFREE | - | + | - | + | + | then LO pulses are as long as HI pulses on the dataline(s) |
| PULSECNTBITS | 5 | 5 | 5 | 4 | 5 | bits for pulse length counter |
| PULSETHRESHOLD | 4 | 9 | 9 | 3 | 3 | number of clock to determine btw. Hi and Lo |
| PULSELONG | 7 | 16 | 16 | 5 | 5 | no of clks for Hi pulse |
| PULSESHORT | 2 | 2 | 3 | 2 | 2 | no of clks for Lo pulse |
| PULSEPAUSE | 3 | 7 | 7 | 1 | 1 | no of clks btw. pulses if not DCFREE |

Table A2.1 FSM variants controlled by Verilog defines in lablet_fsm3_defs.v

## Appendix 3 Simulation of lablets via testbench

### Lablet id0

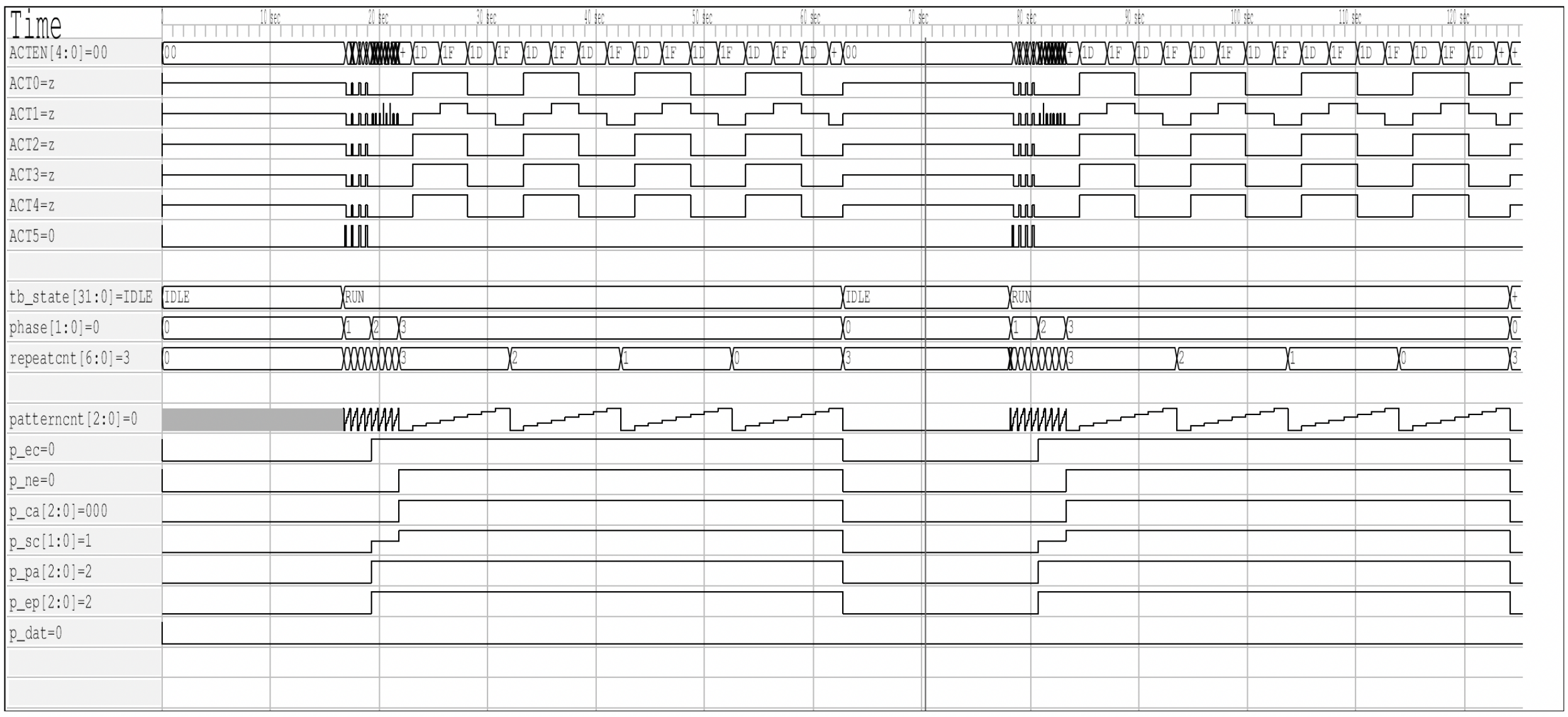

**Figure A3.1 Default program of lablet fsm id0**

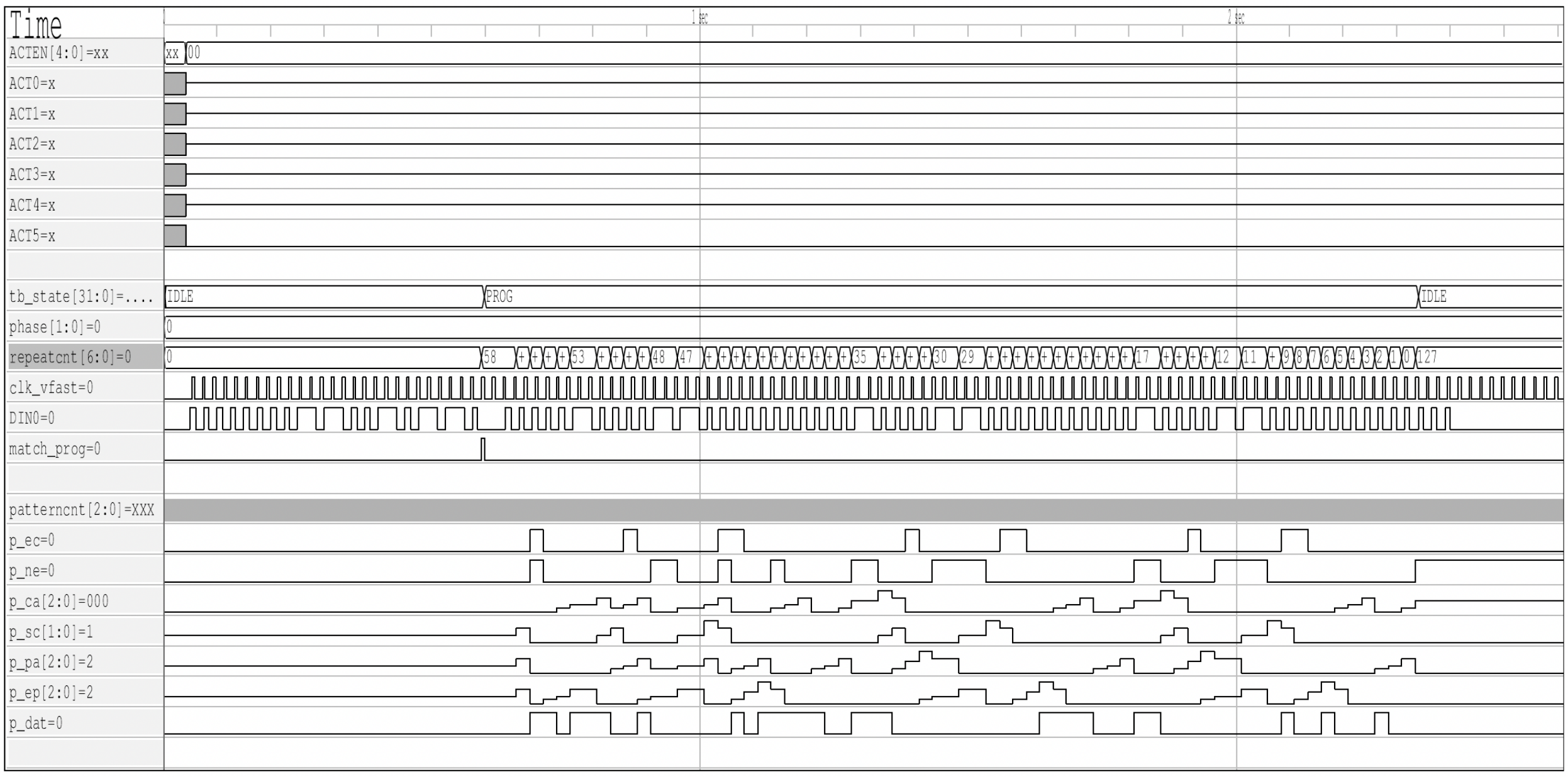

**Figure A3.2 Programming of lablet fsm id0**

a

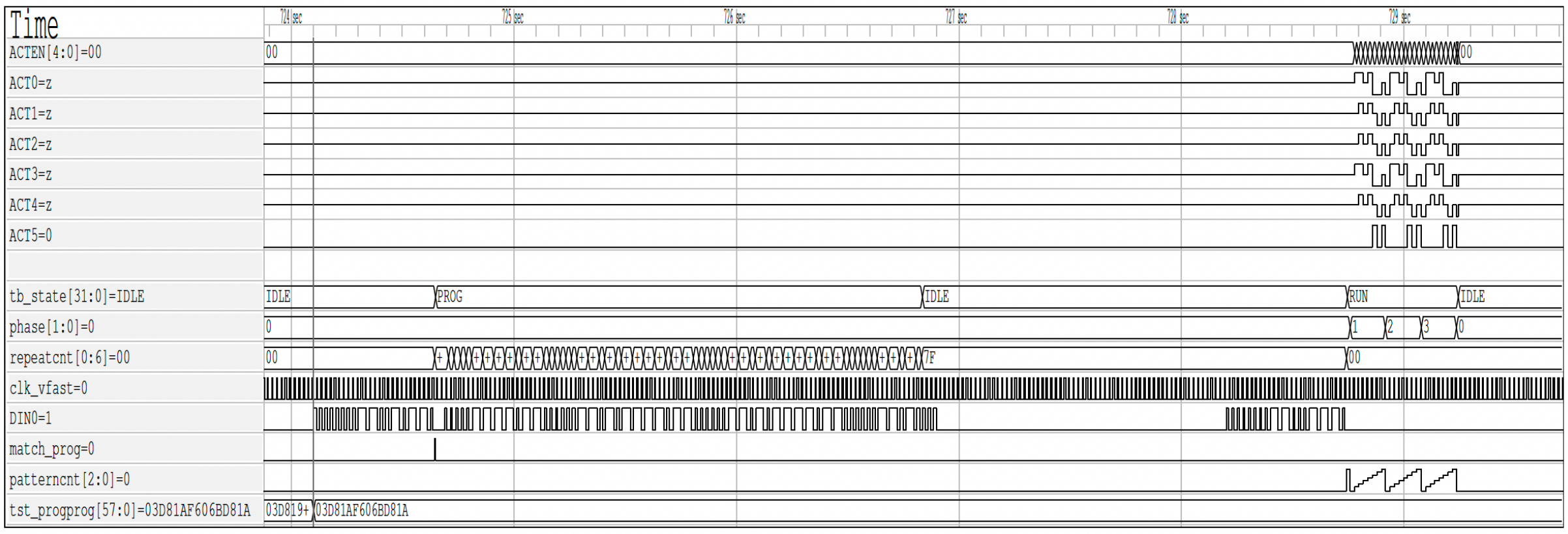

b

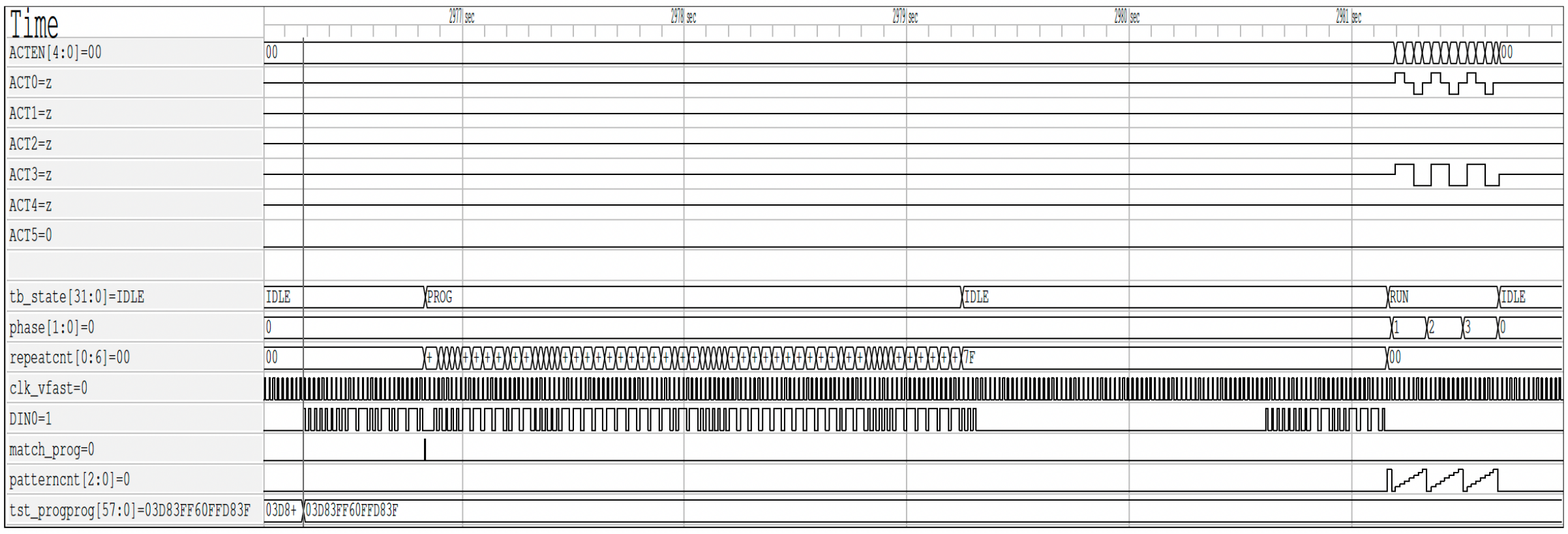

c

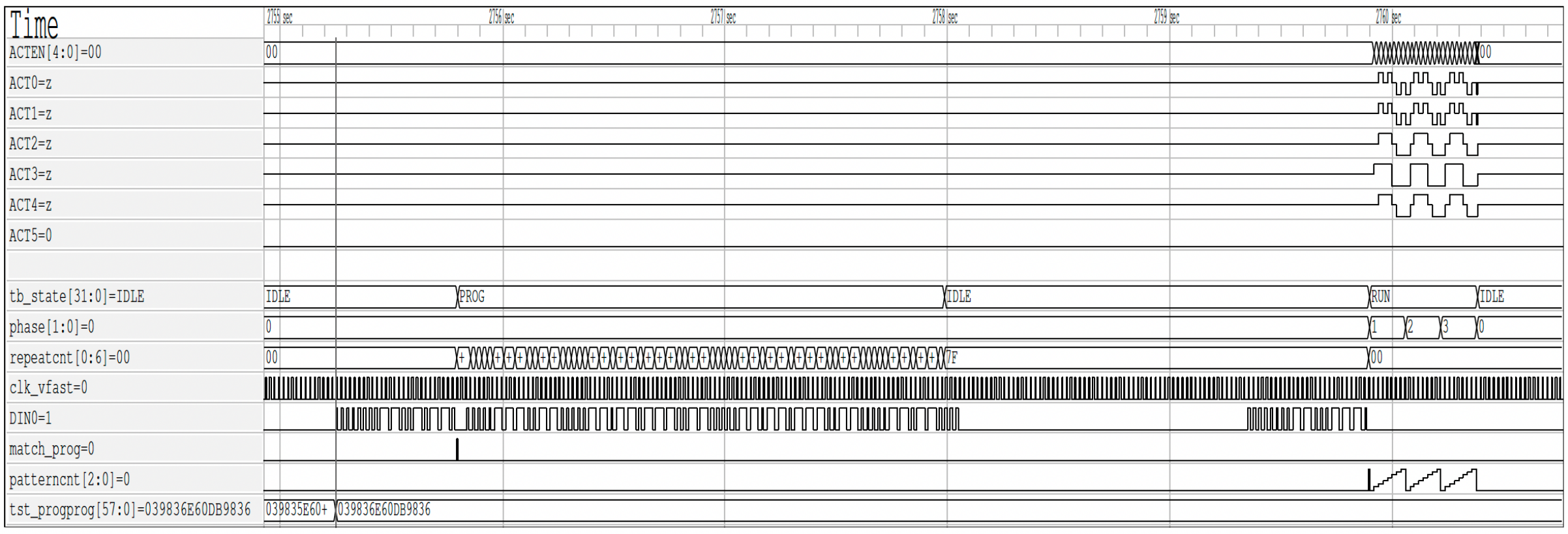

**Figure A3.3 Programming and running lablet id0**. 3 examples showing variety of output waveforms.

## Lablet id1

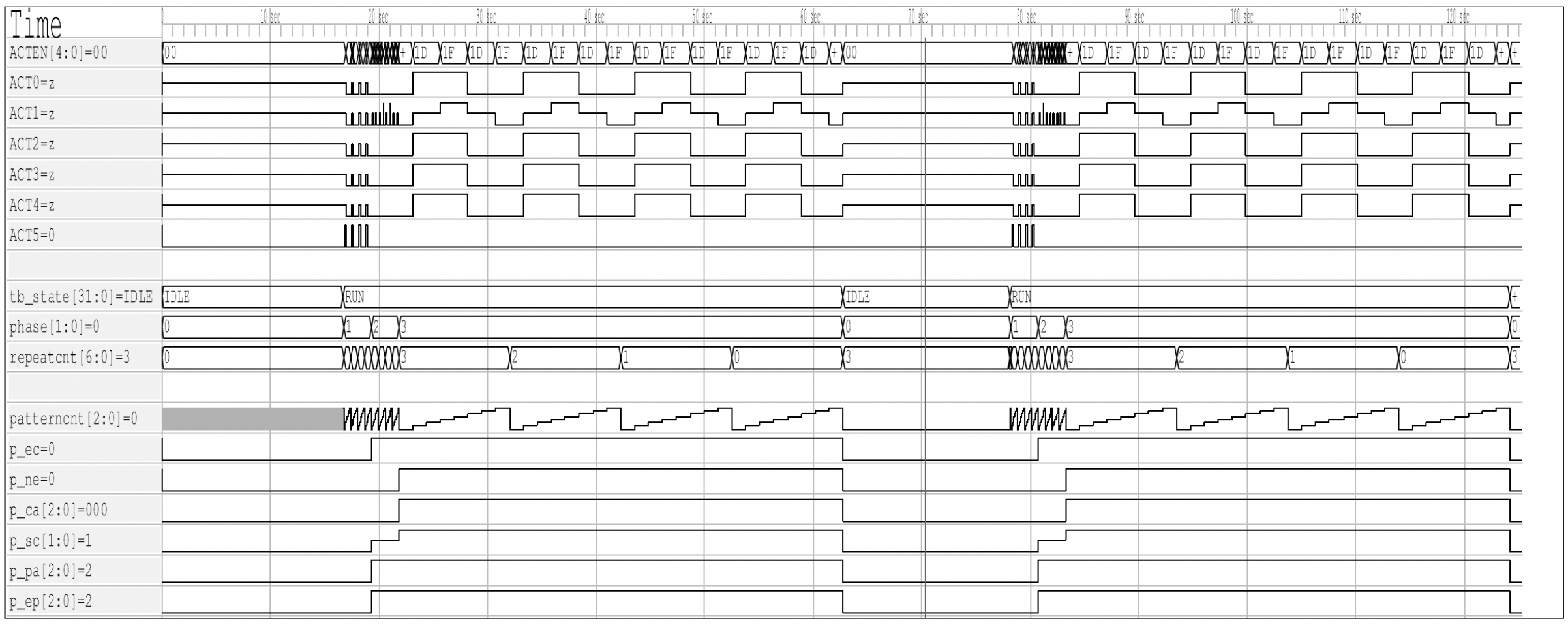

**Figure A3.4 Default program of lablet fsm id1**

**a**

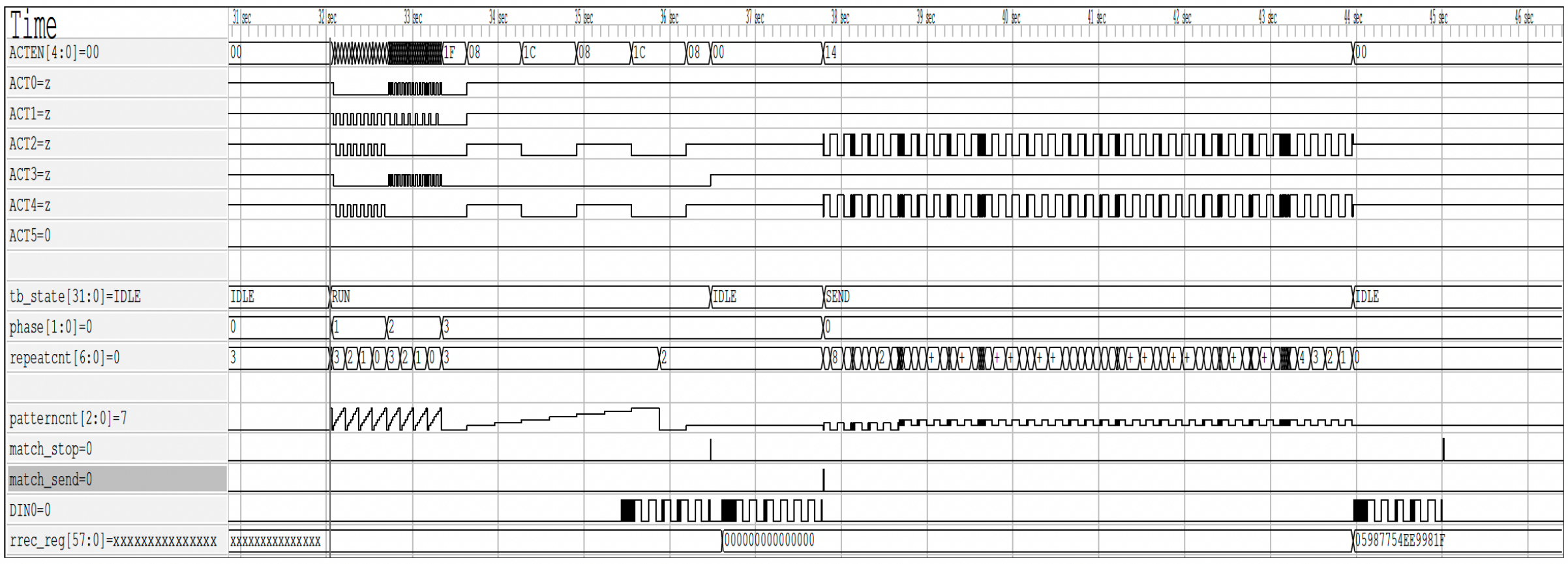

**b**

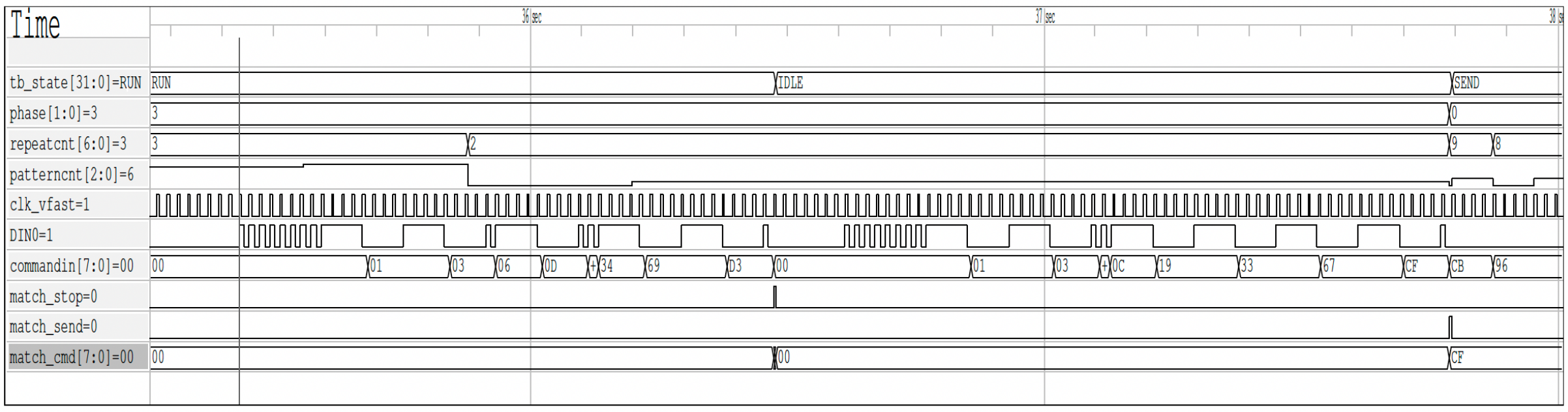

**Figure A3.5 Controlling lablet id1. a** Stopping a running lablet program. **b** Readback program

## Lablet id2

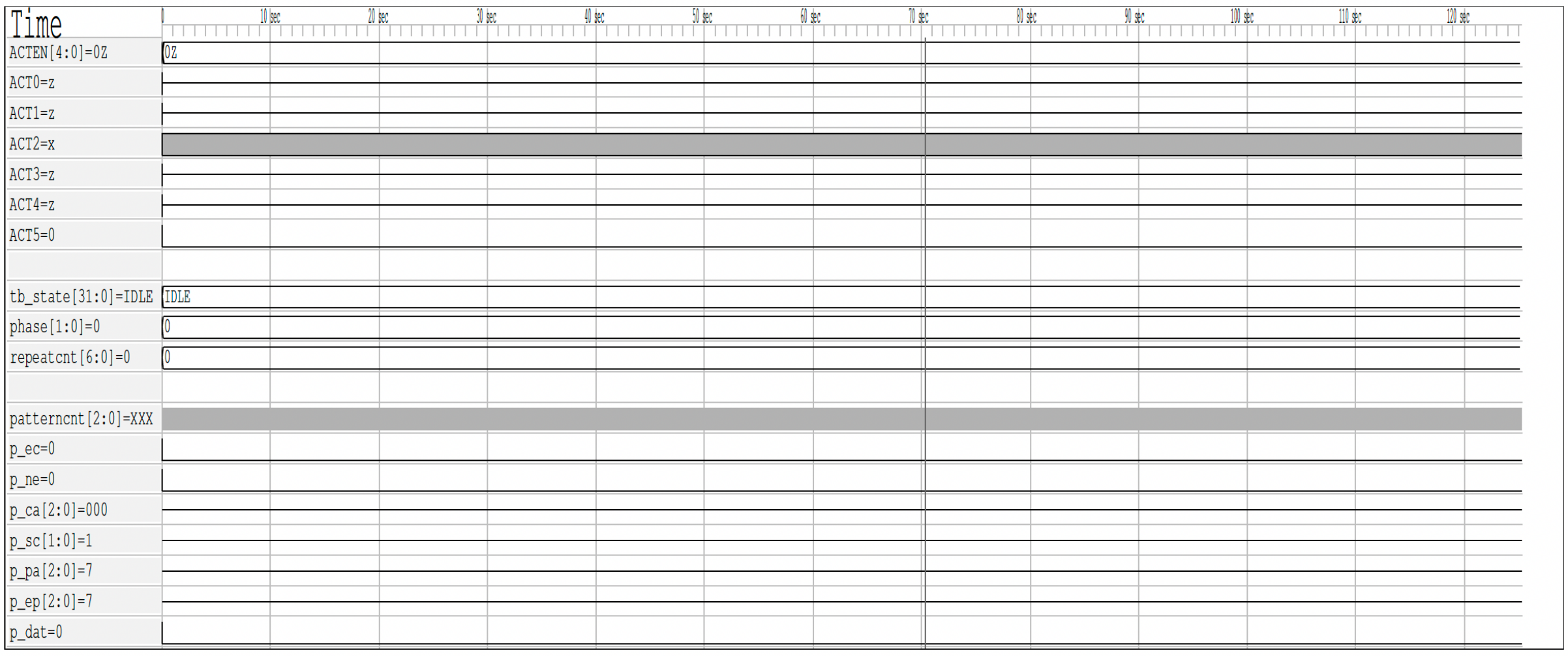

**Figure A3.6 Default program of lablet fsm id2**

## Lablet id3

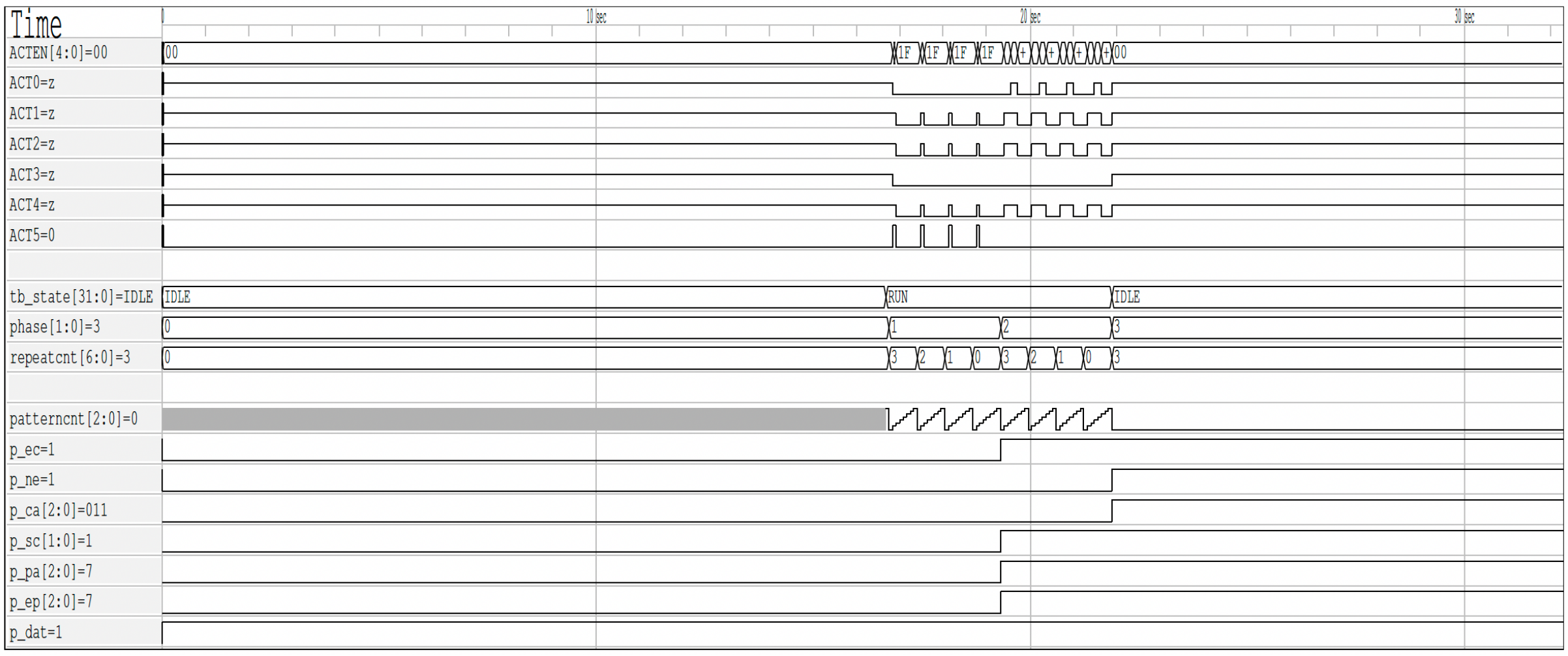

**Figure A3.7 Default program of lablet fsm id3**

## Appendix 4 Lablet array variant keys

The two lablet variant arrays on the CMOS3 reticule or chips cut from it, each with about 1000 different lablets, are encoded graphically in fig. A4.1.

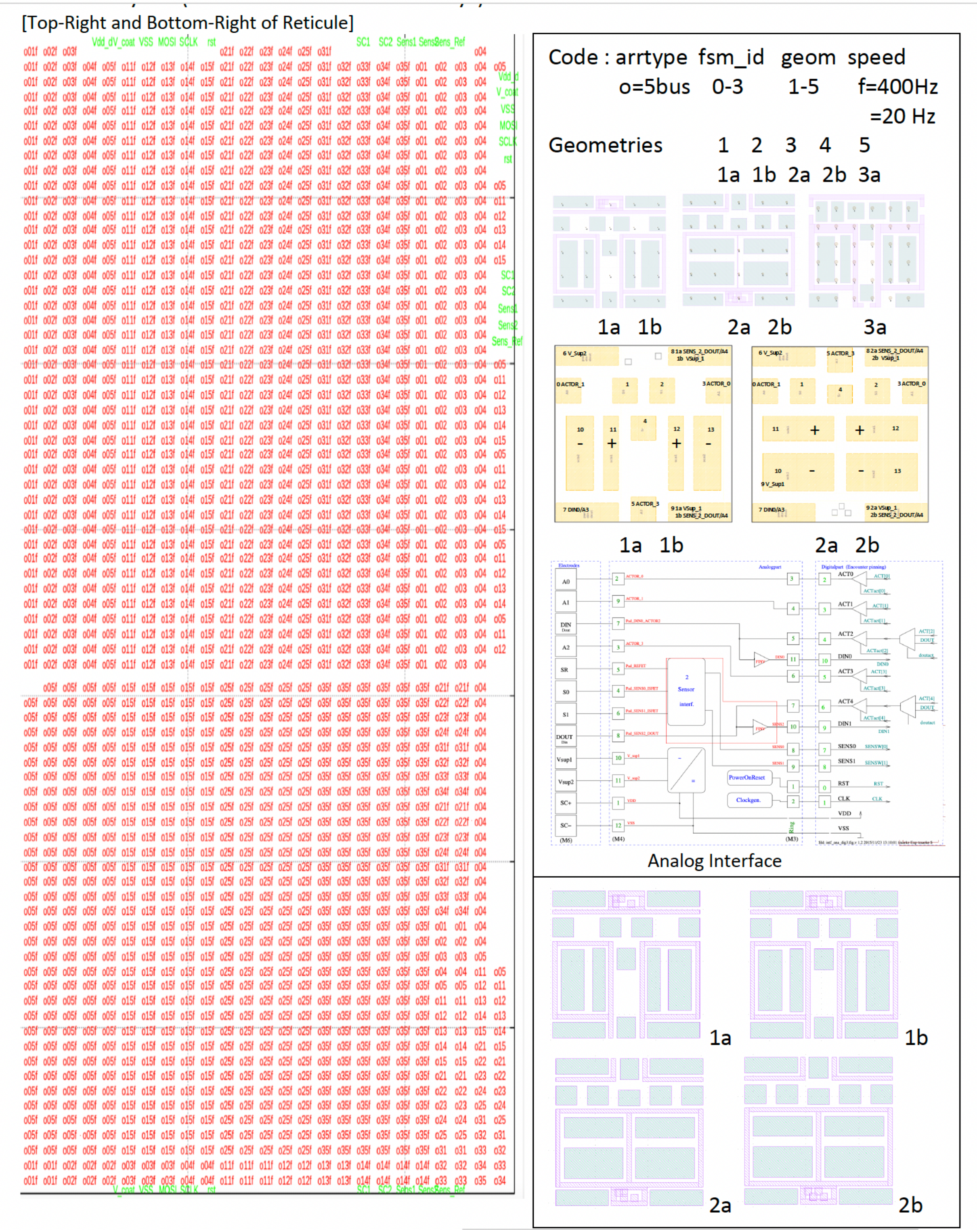

**Fig. A4.1 Codes for variants in two lablet arrays.** 5 geometries, 4 finite state machines and 2 different clock speeds produced combinatorial variants encoded as in the figure.

## Appendix 5 Video sequence of electrochemiluminescence of CMOS3 lablets in array.

The video ECL_CMOS3.mov records the switching of in parallel of $Ru^{2+}(bpy)_3$ electrochemiluminescence (red when voltage exceeds 0.8 V) of actuator electrodes on lablets in a CMOS3 array.